\documentclass[runningheads]{llncs}

\usepackage{eccv}

\usepackage{eccvabbrv}

\usepackage{graphicx}
\usepackage{booktabs}
\usepackage{bbm}
\usepackage{multirow}
\usepackage{scalerel}

\usepackage[accsupp]{axessibility}  

\usepackage{hyperref}

\usepackage{orcidlink}

\begin{document}

\title{Learning to Make Keypoints Sub-Pixel Accurate} 


\author{Shinjeong Kim\inst{1}\orcidlink{0000-0002-3199-4111} \and
Marc Pollefeys\inst{1, 2}\orcidlink{0000-0003-2448-2318} \and
Daniel Barath\inst{1}\orcidlink{0000-0002-8736-0222}}

\authorrunning{S.~Kim et al.}

\institute{Department of Computer Science, ETH Zürich\and
Microsoft Mixed Reality and AI Zurich lab\\
\email{\{shikim, marc.pollefeys, danielbela.barath\}@ethz.ch}}

\maketitle

\begin{abstract}
    This work addresses the challenge of sub-pixel accuracy in detecting 2D local features, a cornerstone problem in computer vision. 
    Despite the advancements brought by neural network-based methods like SuperPoint and ALIKED, these modern approaches lag behind classical ones such as SIFT in keypoint localization accuracy due to their lack of sub-pixel precision. 
    We propose a novel network that enhances \textit{any} detector with sub-pixel precision by learning an offset vector for detected features, thereby eliminating the need for designing specialized sub-pixel accurate detectors. 
    This optimization directly minimizes test-time evaluation metrics like relative pose error. 
    Through extensive testing with both nearest neighbors matching and the recent LightGlue matcher across various real-world datasets, our method consistently outperforms existing methods in accuracy. 
    Moreover, it adds only around 7 ms to the time of a particular detector.
    The code is available at \href{https://github.com/KimSinjeong/keypt2subpx}{https://github.com/KimSinjeong/keypt2subpx}.
\end{abstract}

\section{Introduction}
\label{sec:intro}

The task of identifying and matching sparse 2D feature points across images has long been a cornerstone problem in computer vision~\cite{Harris1988}. 
Algorithms for feature detection enable the generation of detailed 3D models from collections of images~\cite{wu2013towards,jared2015reconstructing,Schonberger2016}, the construction of maps for robotic navigation~\cite{Mur-Artal2015,mur2017orb}, the recognition of places~\cite{schindler2007city,li2010location,noh2017large}, and the estimation of precise locations~\cite{li2012worldwide,toft2018semantic,sattler2016efficient}, as well as facilitating object recognition~\cite{Lowe2004,philbin2007object,nister2006scalable,arandjelovic2012three,arandjelovic2013all}. 
Given its critical importance, the design of algorithms for feature detection and description has attracted considerable attention in computer vision. 
While the seminal SIFT algorithm~\cite{Lowe2004} has been the benchmark for feature detection pipelines for over three decades, the attention has shifted towards learned methods in recent years, with SuperPoint emerging as the gold standard for various applications.

The introduction of advanced machine learning tools has led researchers to substitute traditional, feature-based vision systems with neural networks~\cite{detone2018superpoint,Zhao2022ALIKE,Zhao2023ALIKED}. 
These networks demonstrate increased robustness to variations in features, like changes in viewpoint and illumination, by being explicitly trained on such scenarios with various augmentation techniques. 
However, independent evaluations indicate that these learned models have yet to achieve the keypoint localization precision of their classical predecessors~\cite{Lowe2004}. 
This discrepancy is largely due to the absence of mechanisms for ensuring sub-pixel accuracy\textemdash a fundamental aspect of the long-standing status of the SIFT algorithm as the benchmark in feature detection. 
Additionally, most recent feature detection frameworks design training protocols to mimic the complex conditions faced during the application of feature detectors in vision tasks. Only a select few~\cite{Suwajanakorn2018,bhowmik2020reinforced} are tailored to optimize directly for the specific task they are intended to address.

In this work, we present a network capable of augmenting any learned feature set to deliver sub-pixel precision in keypoint detection. 
The proposed method is structured to append an offset vector to detected features, thereby enabling sub-pixel accuracy without the need for developing entirely new feature detectors. 
This offset is meticulously learned to directly minimize test-time evaluation metrics, for instance, relative pose error.
We demonstrate our improved accuracy through the gold standard SuperPoint~\cite{detone2018superpoint} and the recent ALIKED~\cite{Zhao2023ALIKED} features across several datasets. 
These tests include both nearest neighbors feature matching and the advanced LightGlue matcher~\cite{lindenberger2023lightglue}. 
Results show that our approach consistently improves accuracy metrics in a variety of real-world settings, both indoor and outdoor.

\section{Related Work}

\textbf{Geometrically Invariant Feature Extraction.} Initial efforts in local feature detection and description concentrated on developing carefully engineered algorithms to identify distinctive keypoints and their descriptors, which are robust to variations in viewpoint and illumination. 
Hand-crafted techniques, such as Harris corners~\cite{Harris1988}, SIFT~\cite{Lowe2004}, ORB~\cite{Rublee2011}, and others~\cite{Rosten2006,Calonder2010,Leutenegger2011,Alahi2012,Bay2008,Dalal2005}, utilize explicit geometric concepts like corners, gradients, and scale-space extrema. 
Such approaches became so popular, mainly due to their efficiency and robustness to illumination changes, that they are still crucial components of state-of-the-art frameworks, \eg, for SLAM~\cite{Mur-Artal2015} and Structure-from-Motion~\cite{Schonberger2016}.

Recent advances in deep learning have led to two primary approaches: patch-based descriptor extraction and the integrated learning of keypoints and descriptors. 
Patch-based techniques~\cite{Mishchuk2018,Han2015,Balntas2016,Tian2017,Tian2019}, along with the majority of combined keypoint and descriptor learning strategies~\cite{detone2018superpoint,D2Net2019,DISK2020}, incorporate data augmentation to ensure invariance to scale and orientation changes. 
Certain combined learning approaches go further by explicitly determining the orientation and scale for keypoints. 
LIFT~\cite{Yi2016}, for instance, mimics the SIFT methodology~\cite{Lowe2004} by identifying keypoints, estimating their orientations, and extracting descriptors via separate neural networks. 
It calculates the rotation and scale using a neural network, subsequently applying these transformations to the features to achieve descriptors invariant to orientation and scale changes.

AffNet~\cite{Mishkin2018}, UCN~\cite{Choy2016}, and LF-Net~\cite{Ono2018} advance this approach by estimating affine parameters and applying affine transformations on image features through Spatial Transformer Networks (STN)~\cite{Jaderberg2015}, thereby extracting affine invariant descriptors. 
GIFT~\cite{Liu2019} innovates by creating groups of images at varying scales and orientations, from which it then derives features to produce descriptors that are invariant to scale and orientation. 
Additionally, HDD-Net~\cite{Barroso-Laguna2020} proposes an alternative by rotating convolution kernels instead of the features, aiming to extract descriptors invariant to rotations. REKD~\cite{lee2022self} and RELF~\cite{Lee_2023_CVPR} alternatively utilized a rotation-equivariant convolution layer and devised self-supervised learning frameworks to learn a rotation-equivariant detector and rotation-invariant descriptor, respectively.
%
%

\begin{figure}[tb!]
  \centering
  \includegraphics[width=\linewidth]{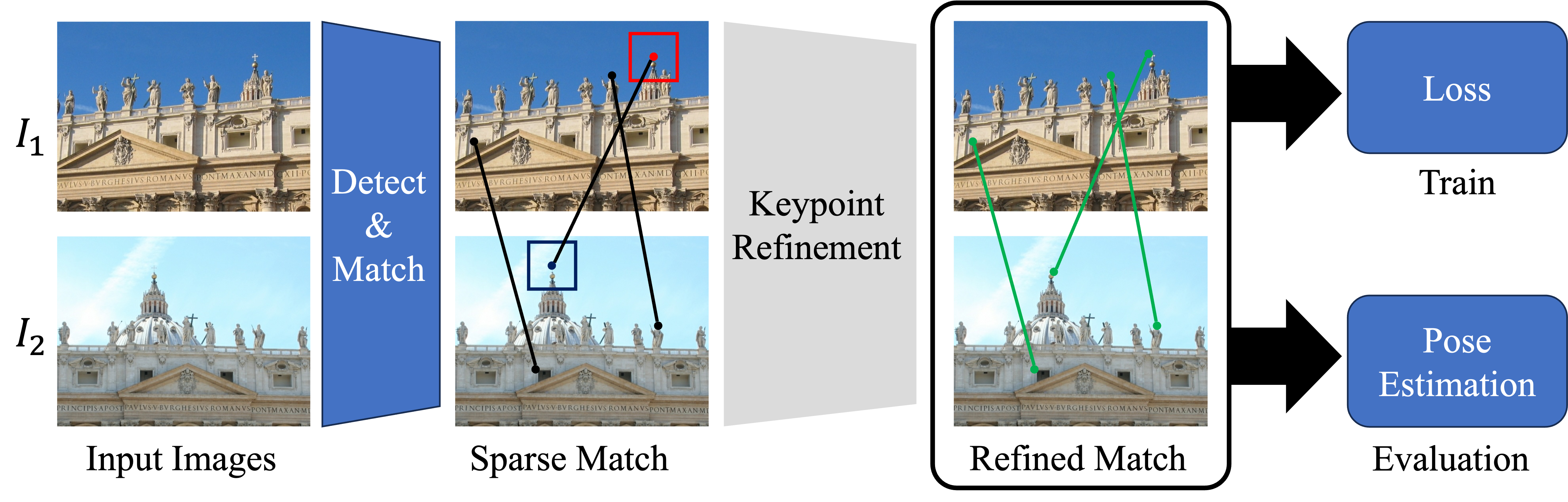}
  \caption{An overview of the proposed sub-pixel refinement method. Given a pair of images, local features are detected, described, and matched to find correspondences. 
  For each match, image patches centered at the keypoints are extracted. Our proposed Keypoint Refinement module (\cref{fig:method}) takes the patches and descriptors to refine the keypoint locations. When training, the refined keypoint matches are used to calculate loss so that the Keypoint Refinement module can be optimized. On evaluation, the relative pose between the two views is estimated using robust estimators.}
  \label{fig:pipeline}
  \vspace{-0.4cm}
\end{figure}

\noindent
\textbf{Joint Keypoint and Descriptor Learning.}
A number of studies have proposed methods to simultaneously estimate both the score map, from which keypoints are detected, and the descriptor map, from which descriptors are sampled. 
SuperPoint~\cite{detone2018superpoint} introduces a lightweight network architecture trained on image pairs augmented automatically through Homographic Adaptation. 
R2D2~\cite{R2D22019} innovates further by calculating repeatability and reliability maps for keypoint detection and employing AP loss for descriptor training. 
A notable enhancement by Suwichaya involves augmenting R2D2 with a Low-Level Feature (LLF) detector to refine keypoint precision~\cite{Suwanwimolkul2021}. 
ALIKE~\cite{Zhao2022ALIKE} distinguishes itself with a differentiable keypoint detection module, enabling precise keypoint training within a highly efficient network suitable for real-time applications. 
Its successor, ALIKED~\cite{Zhao2023ALIKED}, incorporates deformable convolutional networks for enhanced performance. 
While ALIKED obtains sub-pixel accurate features through employing the SoftArgMax operator, it is not explicitly designed to optimize keypoint localization precision.
D2-Net~\cite{D2Net2019}, diverging from score map estimation, identifies keypoints through channel and spatial maxima within the feature map, albeit with a noted limitation in keypoint localization accuracy due to low-resolution feature maps. 
ASLFeat~\cite{Luo2020} leverages multi-level features and deformable convolutions to detect keypoints and model local shapes, thereby improving localization accuracy and descriptor quality. 
D2D~\cite{Tian2020}, drawing inspiration from D2-Net, utilizes a descriptor map alongside absolute and relative saliency for keypoint detection. 
Rao et al.~\cite{Rao2022} introduce a hierarchical view consistency approach to generalize feature descriptors for visual measurements.
While these detectors achieve impressive results across a range of datasets, the keypoint localization accuracy is often poor, as they do not explicitly aim at sub-pixel accuracy.  

\noindent
\textbf{Feature Detection Trained for Specific Tasks.}
The majority of learning-based works in feature detection craft training schemes that replicate the challenging conditions encountered when deploying a feature detector for a specific vision task. 
Contrarily, Key.Net~\cite{Suwajanakorn2018} introduces a differentiable pipeline adept at autonomously identifying category-level keypoints, specifically tailored for the task of relative pose estimation. 
DISK~\cite{DISK2020} utilizes reinforcement learning to train the score and descriptor maps, focusing on the end results of the pipeline to inform its learning process.
Bhowmik et al.~\cite{bhowmik2020reinforced} further integrate feature detection and description within a comprehensive vision pipeline, where the model encounters real-world challenges inherently during training. 
Similarly to DISK, this approach leverages reinforcement learning to address the discrete nature of keypoint selection and descriptor matching.
Its performance is showcased throughout improving the SuperPoint detector~\cite{detone2018superpoint}. 
Additionally, Roessle et al.~\cite{Roessle_2023_ICCV} concentrate on simultaneously improving feature matching and pose optimization by applying differentiable pose estimation techniques.


\section{Sub-Pixel Accurate Local Features}

\begin{figure}[tb!]
  \centering
  \includegraphics[width=\linewidth]{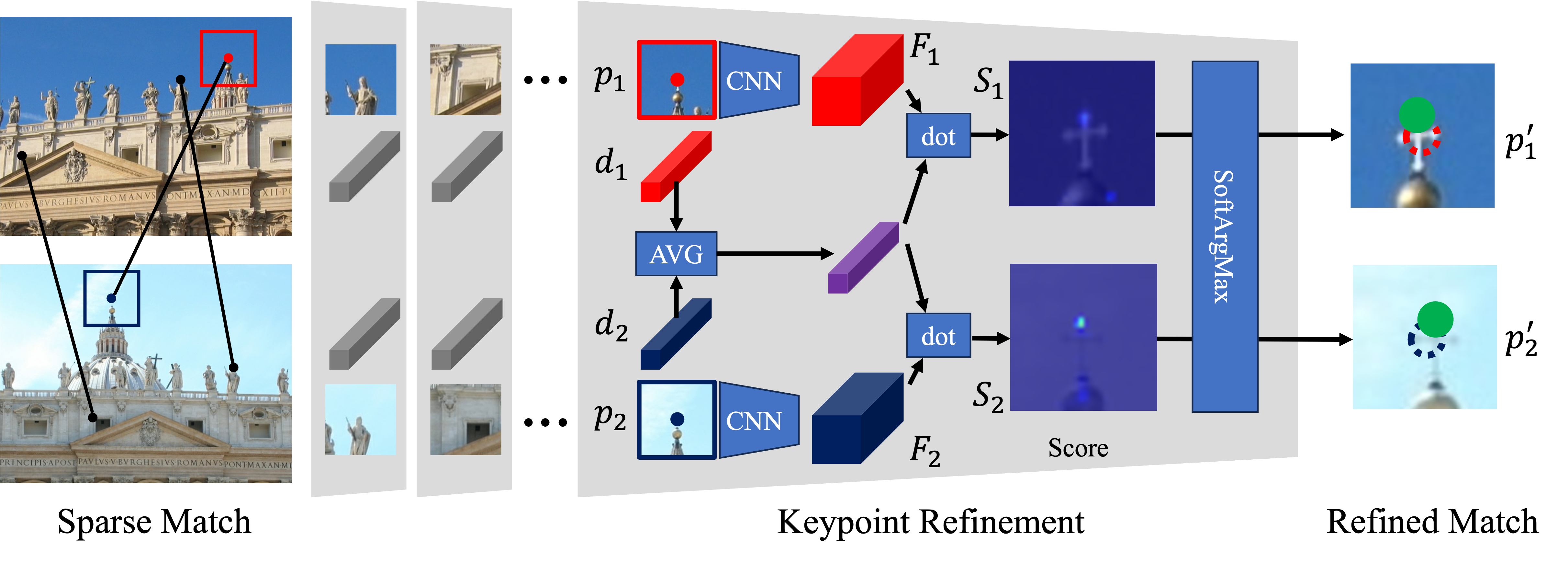}
  \caption{Visualization of how our Keypoint Refinement module works. For detectors producing dense score maps as an intermediate representation, the patches of the score map are concatenated to the image patches.
  The feature map of each patch is extracted with a small convolutional neural network (CNN) and dot-produced with an average of matched descriptor pairs. 
  Taking the SoftArgMax operation on the resulting score map gives the sub-pixel accurate displacement of each keypoint. 
  Note that the weights of two CNNs of the Keypoint Refinement module are shared.}
  \label{fig:method}
  \vspace{-0.4cm}
\end{figure}

We provide a generaliable method to refine any keypoints to be sub-pixel accurate.
%
To this end, we propose a detector-agnostic Keypoint Refinement module guided by local feature descriptors and optimizing a two-view geometric objective directly. 
Our pipeline is designed in a way not only to distill the context information of descriptors to our Keypoint Refinement module but also to consider geometric supervision. 
In this section, we will first introduce our sub-pixel accurate Keypoint Refinement module, followed by the geometric supervision we are providing.

\subsection{Sub-Pixel Accurate Keypoint Refinement}

In this section, we discuss the proposed Keypoint Refinement module to learn an offset vector on top of a detected feature. 
Following de-facto standard process of relative pose estimation based on sparse correspondences, local features are first detected, described, and matched given an image pair. 
Based on the matched feature pairs, we perform our keypoint refinement. 
The brief overview of the proposed architecture is given in \cref{fig:method}.

Next, we describe the refinement procedure for each tentative point correspondence. 
As discussed in several works~\cite{lindenberger2021pixsfm, Zhao2022ALIKE, Zhao2023ALIKED}, local refinement of keypoints can be safely assumed to be independent of the global context. 
Based on the same assumption, we consider a certain size of square windows centering the detected keypoints as our region of interest. 
We extract the region of interest of the image as patches. For the keypoint detectors producing score maps as an intermediate representation, we extract the score patches on the same region and concatenate them with image patches to consider them as side information.

To refine a feature based on the extracted patch centered on the original location of the feature point, we consider the keypoint refinement procedure as predicting new or refining an existing score map. 
To detect keypoints from predicted score maps, either Non-Maximum Suppression (NMS) or its differentiable counterpart, SoftArgMax, are the standard approaches. 
To ensure that the gradient effectively flows through our module, we choose SoftArgMax. 
Given score map $S\in\mathbb{R}^{P\times P}$, SoftArgMax is calculated as follows:
\begin{equation} 
  \mathrm{SoftArgMax}(S) = \sum_{\scaleto{u\in\{-\lfloor \frac{P}{2}\rfloor,\cdots,\lfloor \frac{P}{2}\rfloor\}^2}{6.5pt}}\mathrm{SoftMax}(S)[u]u = \sum_{\scaleto{u\in\{-\lfloor \frac{P}{2}\rfloor,\cdots,\lfloor \frac{P}{2}\rfloor\}^2}{6.5pt}}\frac{e^{S[u]}u}{\sum_{u'}e^{S[u']}}
  \label{eq:softargmax}
\end{equation}
This can be viewed as an expectation of relative keypoint location inside the score patch, where the probability distribution is modeled with SoftMax. 

Our score map prediction procedure is modeled with a Convolutional Neural Network (CNN). 
We extract $D$-dimensional $P\times P$ square-shape normalized features from the image patches, $F_{1i}, F_{2i}\in \mathbb{R}^{P\times P\times D}$ where $F_{1i}$ and $F_{2i}$ correspond to the $i$-th patch of the first image and the second image, respectively. 
We do not assume the convolution layers to have the `same' padding, that is, $P$ does not have to be the dimension of the original image patch used for feature extraction. 
Once we have the feature patches along with corresponding descriptors $d_{1i}, d_{2i}\in \mathbb{R}^D$, we calculate the displacement $\delta_{1i}, \delta_{2i}\in \mathbb{R}^2$ of each keypoint $p_{1i}, p_{2i}\in \{0,\cdots,W\}\times \{0,\cdots,H\}$ in a way to cover the whole input image patch, employing the following equation:
\begin{equation}
  S_{ki}[u]=F_{ki}[u]\cdot \frac{d_{1i}+d_{2i}}{2}\ (k\in\{1, 2\}),
  \label{eq:featuredistill}
\end{equation}
\begin{equation}
  \delta_{ki} = \sigma\times\mathrm{SoftArgMax}(S_{ki}).
  \label{eq:displacement}
\end{equation}
The $\sigma$ is the scale factor to scale the displacement calculated with the feature map back to the scale of the extracted image patch.

The dot product of \cref{eq:featuredistill} guides $F_{1i}$, $F_{2i}$ to learn the embedding space of the descriptor but in pixel-level grained, which is far denser than the initial descriptors. 
The average of two descriptors acts as a regularization on the learned embedding space to be consistent over multiple views, which is a special case of the robust mean of features of PixSfM~\cite{lindenberger2021pixsfm}. The effect of this will be later shown in the experiments. 
The \cref{eq:featuredistill} and \cref{eq:displacement} altogether can also be interpreted as attention with the descriptors as queries, feature patches as keys, and pixel indices inside the patch as values.

The refined keypoints we finally obtain are the sum of the original keypoint and predicted displacement vector as follows:
\begin{equation}
  p'_{1i} = p_{1i} + \delta_{1i}, \quad p'_{2i} = p_{2i} + \delta_{2i},
  \label{eq:refinedkeypt}
\end{equation}
where $p'_{1i}$ and $p'_{2i}$ are the updated keypoint locations in the first and second images, respectively.
During training, gradients only flow toward keypoint displacements ($\delta_{1i}, \delta_{2i}$), as we do not assume the differentiability of the detectors and descriptors we use.
This property makes the proposed method applicable in conjunction with any existing detector.

\subsection{Training Objective}

Our proposed Keypoint Refinement module is directly trained with a geometric objective. 
Here, we explain the formulation and effect of our objective. 
We will conclude this section with why in the previous works, it was difficult to train detectors only with such an objective.

We aim to train the Keypoint Refinement module without any exhaustive annotation on keypoints to enhance the subsequent relative pose estimation. 
While many keypoint detectors utilize consistency under various data augmentation to self-supervise the keypoint locations, our supervision is based on multi-view consistency measured by the epipolar error introduced in the supplementary material of \cite{brachmann2019neural}. 
Given an image pair and its ground truth essential matrix $E$, and matched homogeneous keypoint pairs $\hat{p}_1$ and $\hat{p}_2$, the epipolar error is calculated as follows:
\begin{equation}
  e(\hat{p}_1,\hat{p}_2,E) = \frac{(\hat{p}_2^TE\hat{p}_1)^2}{[E\hat{p}_1]_0^2 + [E\hat{p}_1]_1^2 + [E^T\hat{p}_2]_0^2 + [E^T\hat{p}_2]_1^2},
  \label{eq:epi_err}
\end{equation}
where $[\cdot]_j$ indicates the $j$-th element of the vector inside the bracket.

Directly optimizing the epipolar error can result in an unstable training procedure, as the sum of the errors of all matches is dominated by outliers and thus the keypoints will be optimized toward outliers. To minimize the effect of outliers, we set a threshold $t'$ and only consider errors smaller than that to calculate the loss. Therefore, our epipolar loss is,
\begin{equation}
  \ell(\hat{p}_1,\hat{p}_2,E) = (e(\hat{p}_1,\hat{p}_2,E)\cdot \mathbbm{1}[d(\hat{p}_1,\hat{p}_2,E) < t']) + (t'\cdot\mathbbm{1}[d(\hat{p}_1,\hat{p}_2,E) \geq t']),
  \label{eq:loss}
\end{equation}
where
\begin{equation}
  d(\hat{p}_1,\hat{p}_2,E) = \sqrt{e(\hat{p}_1,\hat{p}_2,E)},
  \label{eq:distance}
\end{equation}
with operator $\mathbbm{1}[\cdot]$ being the Iverson bracket which equals one if the condition inside holds and zero otherwise.

In the calibrated case, we set the threshold $t'$ in a geometrically meaningful way. 
As our choice of robust estimator, GC-RANSAC with $1$ pixel threshold internally uses MSAC score of which margin for inlier selection as $1.5$ times of the threshold, we set $t'$ to correspond to $1.5$ pixel in the normalized image space. 
Following the standard practice, \eg, as done in OpenCV and MAGSAC++~\cite{barath2020magsac++}, we normalize the threshold $t=1.5$ by the focal lengths as follows:
\begin{equation}
 t' = \frac{t}{(f_{1x} + f_{1y} + f_{2x} + f_{2y})/4},
  \label{eq:distance}
\end{equation}
where $f_{ix}$ and $f_{iy}$ are the focal lengths of image $i$ corresponding to the $x$ and $y$ axes, respectively. 

The similar losses based on epipolar geometry and essential matrix are frequently used on optimizing robust estimators\cite{wei2023generalized} or local feature matchers\cite{brachmann2019neural, yi2018learning, zhang2019learning, zhao2021progressive}, however, has yet been applied for keypoint detection.

\begin{table}[tb]
\renewcommand{\arraystretch}{1.15}
  \caption{Relative pose accuracy on datasets MegaDepth~\cite{li2018megadepth}, KITTI~\cite{Geiger2012CVPR} and ScanNet~\cite{dai2017scannet} by running GC-RANSAC~\cite{barath2018graph} on correspondences extracted by vanilla SuperPoint~\cite{detone2018superpoint} (SP), SP refined by the proposed method and by \cite{bhowmik2020reinforced} and matched by Mutual Nearest Neighbor (MNN) matching. 
  We report the AUC scores, calculated from the relative pose errors, thresholded at 5$^\circ$, 10$^\circ$, and 20$^\circ$, 
  the mean and median pose errors in degrees, and
  the inlier ratio of the estimated relative pose. 
  The proposed method consistently outperforms SuperPoint and the reinforcement learning-based approach \cite{bhowmik2020reinforced} in all metrics, except for the mean error on ScanNet, where all methods are similar.}
  \label{tab:spmnn}
  \centering
  \resizebox{\textwidth}{!}{\begin{tabular}{@{}c|l|cccccc@{}}
    \toprule
    Dataset & \multicolumn{1}{c|}{Method} & AUC@$5^{\circ}$ & AUC@$10^{\circ}$ & AUC@$20^{\circ}$ & Inlier Ratio (\%) & Mean ($^{\circ}$) & Median ($^{\circ}$)\\
    \midrule
    \multirow{3}{*}{MegaDepth} & SP~\cite{detone2018superpoint} + MNN & 35.34 & 45.37 & 54.22 & 33.78 & 27.89 & 3.94\\
    & Reinforced SP~\cite{bhowmik2020reinforced} + MNN & 36.61 & 47.56 & 56.71 & 33.90 & 27.91 & 3.92\\
    & SP + MNN + Ours & \textbf{37.16} & \textbf{48.07} & \textbf{57.15} & \textbf{34.42} & \textbf{27.69} & \textbf{3.82} \\
    \midrule
    \multirow{3}{*}{KITTI} & SP~\cite{detone2018superpoint} + MNN & 80.92 & 89.49 & 94.41 & 73.27 & 1.20 & 0.62\\
    & Reinforced SP~\cite{bhowmik2020reinforced} + MNN & 80.97 & 89.53 & 94.43 & 73.27 & 1.20 & 0.62\\
    & SP + MNN + Ours & \textbf{81.36} & \textbf{89.69} & \textbf{94.49} & \textbf{73.47} & \textbf{1.16} & \textbf{0.61} \\
    \midrule
    \multirow{3}{*}{ScanNet} & SP~\cite{detone2018superpoint} + MNN & 13.92 & 28.16 & 43.11 & 37.56 & \textbf{28.58} & 9.76\\
    & Reinforced SP~\cite{bhowmik2020reinforced} + MNN & 13.83 & 28.11 & 43.17 & 37.55 & 28.60 & 9.76\\
    & SP + MNN + Ours & \textbf{14.16} & \textbf{28.57} & \textbf{43.56} & \textbf{37.66} & 28.68 & \textbf{9.59}\\
  \bottomrule
  \end{tabular}}
\end{table}

The challenge of achieving differentiation has been a significant barrier to enhancing feature detection methods with sub-pixel precision. 
Previous attempts struggled with differentiating components of the vision pipeline, such as detectors like SuperPoint~\cite{detone2018superpoint} or matchers like Nearest Neighbor Matching. 
These efforts often encountered unstable training due to the complexities of integrating detectors and matches. 
Successful differentiation in prior work was either achieved by forgoing end-to-end differentiability through the use of the REINFORCE algorithm~\cite{bhowmik2020reinforced} or was only demonstrated in simple, driving scenarios~\cite{2020_jau_zhu_deepFEPE}. 
Our approach overcomes these issues by integrating our Keypoint Refinement module directly before the relative pose estimation stage of the pipeline, facilitating stable and effective differentiation.

\begin{table}[tb]
    \renewcommand{\arraystretch}{1.15}
  \caption{Relative pose accuracy on datasets MegaDepth~\cite{li2018megadepth}, KITTI~\cite{Geiger2012CVPR} and ScanNet~\cite{dai2017scannet} by running GC-RANSAC~\cite{barath2018graph} on SuperPoint~\cite{detone2018superpoint} or ALIKED~\cite{Zhao2023ALIKED} features matched by LightGlue~\cite{lindenberger2023lightglue} and refined by the proposed method.
  We report the AUC scores, calculated from the relative pose errors, thresholded at 5$^\circ$, 10$^\circ$, and 20$^\circ$, 
  the mean and median pose errors in degrees, and
  the inlier ratio of the estimated relative pose. 
  The proposed method consistently improves both detectors in almost all metrics.}
  \label{tab:splgresults}
  \centering
  \resizebox{\textwidth}{!}{\begin{tabular}{@{}c|l|cccccc@{}}
    \toprule
    Dataset & \multicolumn{1}{c|}{Method} & AUC@$5^{\circ}$ & AUC@$10^{\circ}$ & AUC@$20^{\circ}$ & Inlier Ratio (\%) & Mean ($^{\circ}$) & Median ($^{\circ}$)\\
    \midrule
     \multirow{4}{*}{MegaDepth} & ALIKED+LG & 64.69 & 76.77 & 85.17 & 87.10 & \phantom{1}6.65 & 1.01\\
     & ALIKED+LG+Ours & \textbf{65.34} & \textbf{77.27} & \textbf{85.52} & \textbf{87.73} & \phantom{1}\textbf{6.57} & \textbf{0.98}\\
     \cmidrule{2-8}
     & SP+LG & 59.82 & 72.88 & 82.31 & 79.34 & \phantom{1}7.39 & 1.25\\
     & SP+LG+Ours & \textbf{61.82} & \textbf{74.46} & \textbf{83.41} & \textbf{81.24} & \phantom{1}\textbf{7.22} & \textbf{1.15}\\
     \midrule
    \multirow{4}{*}{KITTI} & ALIKED+LG & 84.24 & 91.45 & \textbf{95.36} & 79.88 & \phantom{1}1.24 & \textbf{0.50}\\
     & ALIKED+LG+Ours & \textbf{84.27} & \textbf{91.46} & \textbf{95.36} & \textbf{80.02} & \phantom{1}\textbf{1.20} & \textbf{0.50}\\
     \cmidrule{2-8}
     & SP+LG & 81.50 & 89.68 & 94.37 & 81.28 & \phantom{1}1.40 & 0.59\\
     & SP+LG+Ours & \textbf{81.99} & \textbf{89.95} & \textbf{94.51} & \textbf{81.59} & \phantom{1}\textbf{1.34} & \textbf{0.57}\\
     \midrule
    \multirow{4}{*}{ScanNet} & ALIKED+LG & 20.99 & 38.68 & 54.81 & 70.92 & \textbf{20.35} & 5.86\\
     & ALIKED+LG+Ours & \textbf{21.21} & \textbf{38.90} & \textbf{55.00} & \textbf{70.98} & {20.39} & \textbf{5.82}\\
     \cmidrule{2-8}
     & SP+LG & 19.40 & 37.96 & 55.46 & 85.51 & 20.07 & 5.97\\
     & SP+LG+Ours & \textbf{19.71} & \textbf{38.26} & \textbf{55.80} & \textbf{85.67} & \textbf{19.84} & \textbf{5.89} \\
  \bottomrule
  \end{tabular}}
\end{table}

\section{Implementation Details}
\textbf{Network Architecture Details. }
Our Convolutional Neural Network responsible for refining (or producing, when used along with detectors involving no heatmaps) the score heatmap consists of a series of convolutional layers with $3\times 3$ sized filters followed by ReLU activations, without any extra poolings in between. 
The numbers of the channel sizes are 16, 16, 64, 64, $d$, where $d$ is the dimension of the descriptors. 
The `same' padding is only applied on layers where the input and output channel sizes are the same. Thus, the dimension of the feature map starts from 11$\times$11 and is eventually reduced to 5$\times$5, which covers distances of $2$ pixel toward each direction from the original keypoints. To scale this amount back to cover the entire region of the original patch, $\sigma$ value for displacement scaling is set to $2.5$, since $\frac{\lfloor 11/2\rfloor}{\lfloor 5/2\rfloor} = 2.5$. Thus, the resulting displacement vector covers distances of $2\times2.5 = 5$ pixel toward each direction. 
This enables our model to confine its refined keypoints inside its input image patch, and thus the refined keypoints do not diverge too far away from their original location, stabilizing the training of our module.
The implementation of our model is based on PyTorch, taking only 123K (ALIKED) and 197K (SuperPoint) more parameters than the original detector.

\noindent\textbf{Training Details.} We train our networks with Adam optimizer with the default hyperparameters of PyTorch 1.10, except the learning rate which is set to $1\times10^{-4}$. Our pipeline is trained for 200K steps with a batch size of 8, taking approximately 12 hours with a decent modern GPU (NVIDIA RTX 3080Ti), which is also used for all evaluations.
Our training is done on the training set of the MegaDepth~\cite{li2018megadepth} dataset.
We use this model\textemdash trained for each detector\textemdash in \textit{all} experiments, indoors and outdoors.

\noindent\textbf{Evaluation Details.} Our choice of robust estimator to estimate essential matrices is the state-of-the-art GC-RANSAC~\cite{barath2018graph}. 
The official implementation of GC-RANSAC runs an inlier resampling-based local optimization~\cite{chum2003locally} and ends with a Levenberg-Marquardt~\cite{more2006levenberg}.
It employs PROSAC sampling~\cite{chum2005matching} and the SPRT test~\cite{chum2008optimal} to achieve efficiency.
We set the minimum and maximum iteration number to 1000 to minimize the effect of stochasticity and ran each experiment 3 times to report the average. 
We set the threshold of GC-RANSAC 1 px.

We evaluate the proposed sub-pixel refinement method in conjunction with the gold standard SuperPoint~\cite{detone2018superpoint} and the recent ALIKED~\cite{Zhao2023ALIKED} detectors, with mutual nearest neighbors (MNN) and LightGlue~\cite{lindenberger2023lightglue} matching. 
We will compare to the original detectors and also to the reinforcement learning-based method proposed by Bhowmik et al.~\cite{bhowmik2020reinforced}.

\noindent\textbf{Evaluation Metrics.}
To assess the performance of a particular method, we follow the standard practice~\cite{IMC2020} and measure the relative pose error as the maximum of the rotation and translation errors. 
From these numbers, we will compute the Area Under the recall Curves (AUC) thresholded at 5$^\circ$, 10$^\circ$, and 20$^\circ$.
Additionally, we will report the mean, median of pose error, and inlier ratios. 

\vspace{-0.15cm}
\section{Experiments} 

In this section, the evaluation is conducted on two outdoor datasets (MegaDepth and KITTI~\cite{Geiger2012CVPR}) and one indoor dataset (ScanNet~\cite{dai2017scannet}). 
Additionally, we provide computational analysis and an analysis of the proposed method.

\begin{table}[tb]
    \renewcommand{\arraystretch}{1.15}
  \caption{Fundamental matrix accuracy on datasets MegaDepth~\cite{li2018megadepth}, KITTI~\cite{Geiger2012CVPR} and ScanNet~\cite{dai2017scannet} by running GC-RANSAC~\cite{barath2018graph} on correspondences extracted by vanilla SuperPoint~\cite{detone2018superpoint} (SP), SP refined by the proposed method and by \cite{bhowmik2020reinforced} and matched by Mutual Nearest Neighbor (MNN) matching. 
  We report the F1 scores, the inlier ratio, and the mean and median epipolar errors given the ground truth fundamental matrix. 
  The proposed method consistently outperforms SuperPoint and the reinforcement learning-based approach \cite{bhowmik2020reinforced} in all metrics, except for the mean error on ScanNet.}
  \centering
  \resizebox{\textwidth}{!}{\begin{tabular}{@{}c|l|cccccc@{}}
    \toprule
    Dataset & \multicolumn{1}{c|}{Method} & F1-score & Epi. Inlier Ratio (\%) & Epi. Mean & Epi. Med \\
    \midrule
    \multirow{3}{*}{MegaDepth} & SP~\cite{detone2018superpoint} + MNN & 47.64 & 18.64 & \textbf{84.15} & 1.69 \\
    & Reinforced SP~\cite{bhowmik2020reinforced} + MNN & 47.62 & 18.65 & 84.32 & 1.70\\
    & SP + MNN + Ours & \textbf{48.42} & \textbf{19.64} & 84.41 & \textbf{1.67} \\
    \midrule
    \multirow{3}{*}{KITTI} & SP~\cite{detone2018superpoint} + MNN & 74.15 & 48.48 & \phantom{1}0.91 & 0.81 \\
    & Reinforced SP~\cite{bhowmik2020reinforced} + MNN & 74.12 & 48.47 & \phantom{1}0.91 & 0.81 \\
    & SP + MNN + Ours & \textbf{75.48} & \textbf{51.82} & \phantom{1}\textbf{0.87} & \textbf{0.77} \\
    \midrule
    \multirow{3}{*}{ScanNet} & SP~\cite{detone2018superpoint} + MNN  & 18.83 & 15.83 & \textbf{45.86} & 4.96\\
    & Reinforced SP~\cite{bhowmik2020reinforced} + MNN & 18.76 & 15.82 & 46.42 & 4.95 \\
    & SP + MNN + Ours & \textbf{18.91} & \textbf{16.23} & 47.11 & \textbf{4.83}\\
  \bottomrule
  \end{tabular}}
  \label{tab:fund}
  \vspace{-0.4cm}
\end{table}

\vspace{-0.2cm}
\subsection{Relative Pose Estimation}

To perform relative pose estimation, we run GC-RANSAC to estimate essential matrices.
These matrices are then decomposed to relative rotation and translation by the standard SVD-based solution~\cite{hartley2003multiple}.

\noindent
The \textbf{MegaDepth}
 dataset~\cite{li2018megadepth} is derived from a diverse collection of internet-sourced images depicting landmarks across the globe.
This dataset leverages sophisticated structure-from-motion (SfM) and multi-view stereo (MVS) techniques to generate detailed reconstructions and depth maps for over 1000 distinct scenes. 
We utilize the benchmark consisting of 1500 image pairs, following the selection of LoFTR~\cite{sun2021loftr}, as the validation set and the test set consisting of 655K image pairs coupled by~\cite{lindenberger2023lightglue, pautrat_suarez_2023_gluestick} as our test set.

The results when using SuperPoint features and mutual nearest neighbors matching are reported in the top three rows of Table~\ref{tab:spmnn}. 
The proposed refinement method consistently improves in all accuracy metrics. 
It improves upon the vanilla SuperPoint, by approximately $2$ AUC points on average. 
It also increases the inlier ratio and decreases both the mean and median errors.
We also outperform the previous refinement method~\cite{bhowmik2020reinforced} in all evaluated metrics.

The results with SuperPoint and ALIKED feature when employing the recent LightGlue matcher are shown in the top rows of Table~\ref{tab:splgresults}. 
Similarly to Table~\ref{tab:spmnn}, the proposed method improves upon both ALIKED and SuperPoint.
While the accuracy increment is sometimes small, it is consistent\textemdash we improve in all tested accuracy metrics.

\noindent
The \textbf{KITTI} dataset~\cite{Geiger2012CVPR} is a real-world benchmark for tasks stereo, optical flow, visual odometry, 3D object detection, and tracking. 
It is captured by driving in the city of Karlsruhe with accurate ground truth from the laser scanner and GPS localization system. 
We tested our method on the 11 visual odometry sequences that are provided with ground truth. 
The KITTI experiments are based on the odometry test set with 2790 image pairs defined in \cite{2020_jau_zhu_deepFEPE}.
We form image pairs from the subsequent images. 

The results of SuperPoint and MNN matching are shown in the middle three rows of Table~\ref{tab:spmnn}. 
Similar to the results on the MegaDepth dataset, the proposed method consistently improves in all accuracy metrics. 
It improves upon the vanilla SuperPoint in AUC points, increases the inlier ratio, and decreases both the mean and median errors.
We outperform a previous work that improved SuperPoint with a relative pose supervision~\cite{bhowmik2020reinforced} in all evaluated metrics.

The results with SuperPoint and ALIKED features when employing the recent LightGlue matcher are shown in the middle of Table~\ref{tab:splgresults}. 
Similarly to Table~\ref{tab:spmnn}, the proposed method improves upon both ALIKED and SuperPoint.
A similar trend is as observable as on the other datasets. The proposed technique always improves accuracy, not by a huge margin, but consistently across all metrics.

\noindent
The \textbf{ScanNet} dataset~\cite{dai2017scannet} contains 1613 monocular sequences with ground truth camera poses and depth maps.
We evaluate the compared minimal solvers on the 1500 challenging pairs used in SuperGlue~\cite{sarlin2020superglue}.
The results in the bottom three rows of Table~\ref{tab:spmnn} exhibit similar trends as on the other datasets.
We consistently improve in almost all metrics. 
Here, the mean error increases marginally. However, all methods achieve similar mean errors. 

The efficacy of our proposed method is further evidenced in the bottom section of Table~\ref{tab:splgresults}, where results using SuperPoint and ALIKED features in conjunction with the recent LightGlue matcher are presented. Mirroring the improvements seen in Table~\ref{tab:spmnn}, our method enhances the performance of both ALIKED and SuperPoint. 
This consistency aligns with trends observed on other datasets, where our technique uniformly bolsters accuracy. 
While the increments in performance are not huge, they are reliably observed across all evaluated metrics, underscoring the method's consistent contribution to accuracy enhancement.


\begin{figure}[t!]
    \begin{subfigure}{0.32\textwidth}  
        \centering
        \includegraphics[width=\linewidth]{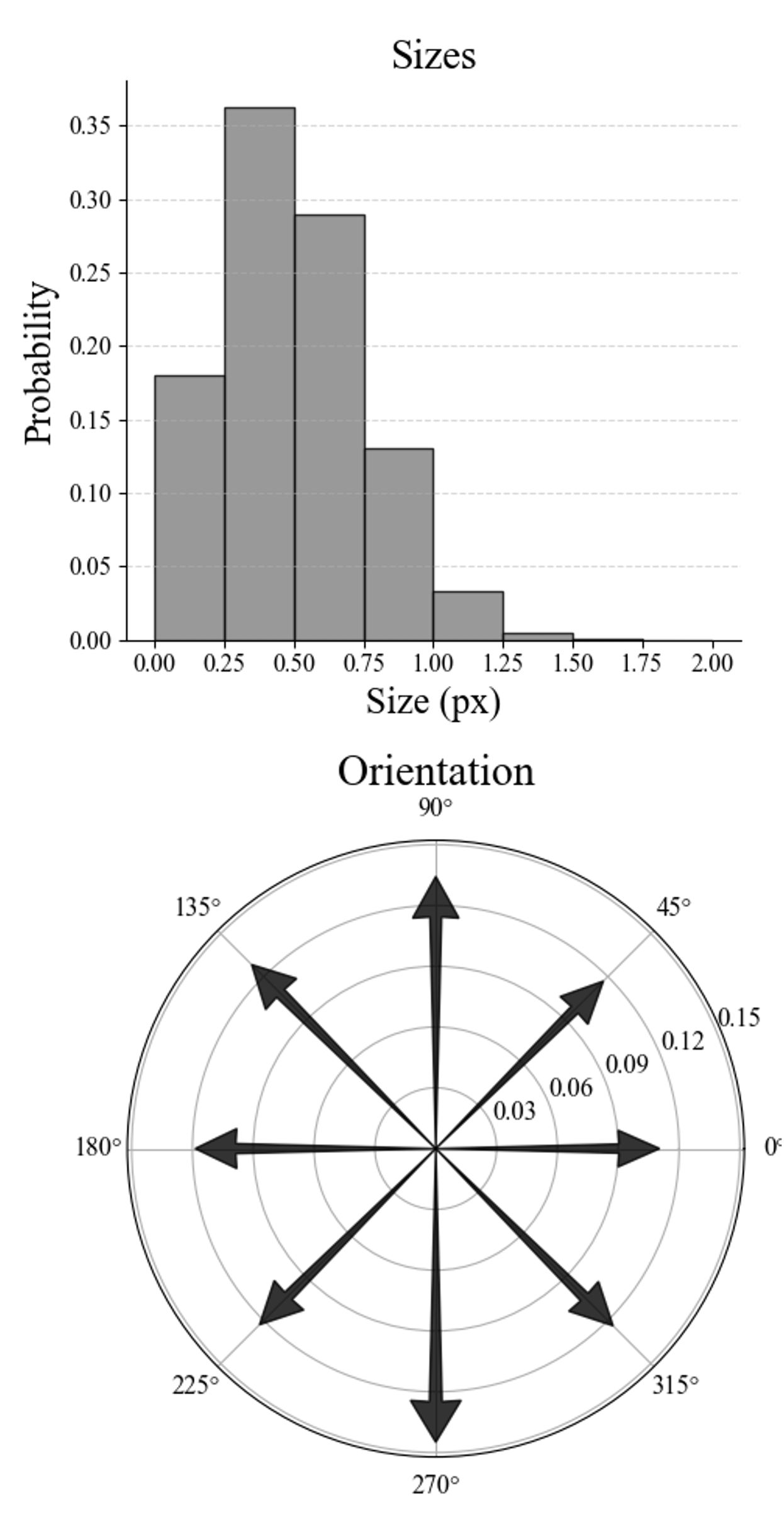}
        \caption{MegaDepth}
    \end{subfigure}
    \begin{subfigure}{0.32\textwidth}  
        \centering
        \includegraphics[width=\linewidth]{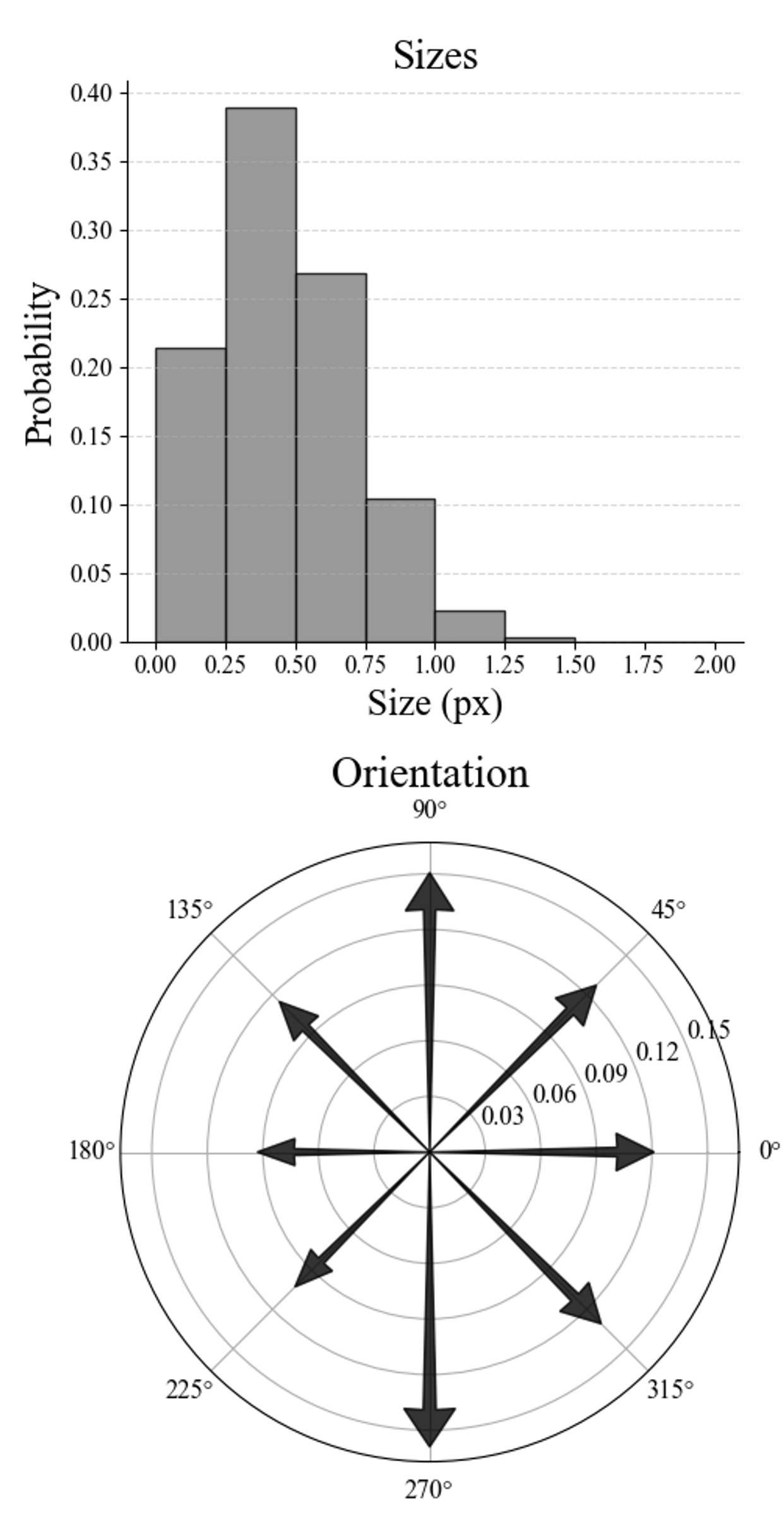}
        \caption{KITTI}
    \end{subfigure}
    \begin{subfigure}{0.32\textwidth}  
        \centering
        \includegraphics[width=\linewidth]{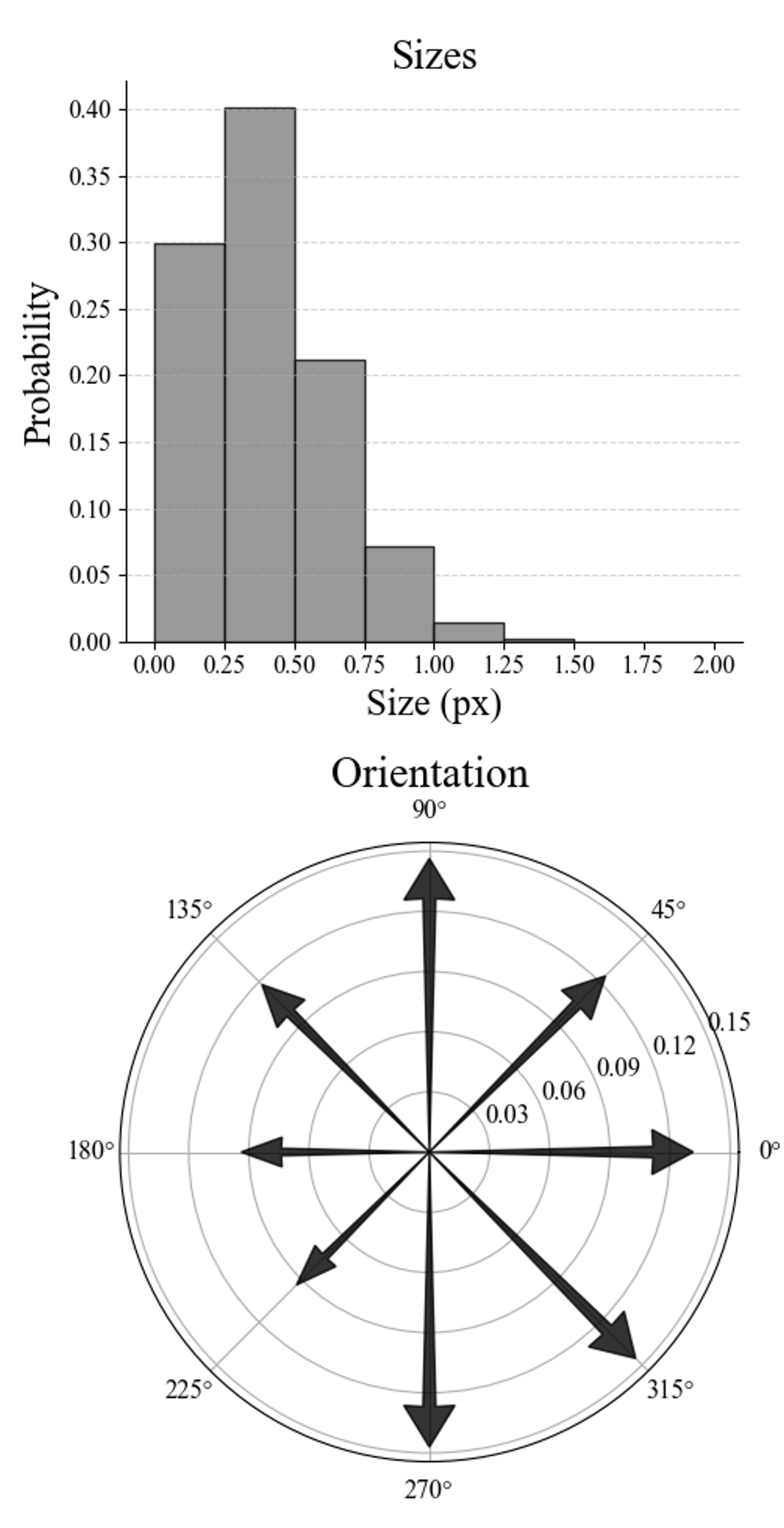}
        \caption{ScanNet}
    \end{subfigure}
    \caption{Histograms of scale and orientation of offset vectors for SuperPoint across various datasets. 
    The top row illustrates the distribution of lengths of offset vectors, with significant displacements observed on MegaDepth, suggesting a high potential for accuracy improvements through sub-pixel refinement due to initial keypoint localization inaccuracies. 
    The progression from left to right shows decreased lengths, with minimal adjustments required on ScanNet. 
    The bottom row displays the directional histogram of predicted offset vectors, revealing a uniform distribution with a tendency for vertical alignment and a rightward bias, likely reflecting unique characteristics of datasets. We do not show the bins beyond offset size $>2$px as they are negligible}
    \label{fig:histograms}
  \vspace{-0.3cm}
\end{figure}

\subsection{Fundamental Matrix Estimation}

In this section, we extend the evaluation of our proposed method to assess its impact on fundamental matrix estimation, a critical task when camera intrinsics are unknown. 
Utilizing the identical datasets and experimental setup as outlined in the previous section, we employ GC-RANSAC~\cite{barath2018graph} to derive fundamental matrices from the detected features. 
The assessment focuses on F1 scores, average inlier ratios, and both mean and median symmetric epipolar errors in relation to the ground truth fundamental matrix.

Findings, as detailed in Table~\ref{tab:fund}, echo the enhancements observed in the relative pose accuracy experiments. 
The integration of the sub-pixel refinement mechanism with vanilla SuperPoint consistently improves performance across all evaluated accuracy metrics. 
An exception to this trend is noted in the mean error metric within the ScanNet dataset. 
It is pertinent to acknowledge that the mean error, while informative, is often regarded as less critical compared to other metrics due to its susceptibility to distortion by outliers; a single significant error can disproportionately affect the mean. 
Notably, in these experiments, the inclusion of reinforced SuperPoint~\cite{bhowmik2020reinforced} results in a decrement in accuracy. 
This observation underscores the nuanced effectiveness of sub-pixel refinement in enhancing feature-based estimations of fundamental matrices, highlighting the proposed method's broad applicability and potential for improving feature detection and matching processes.




\begin{figure}[tb]
  \centering
  \includegraphics[width=\linewidth]{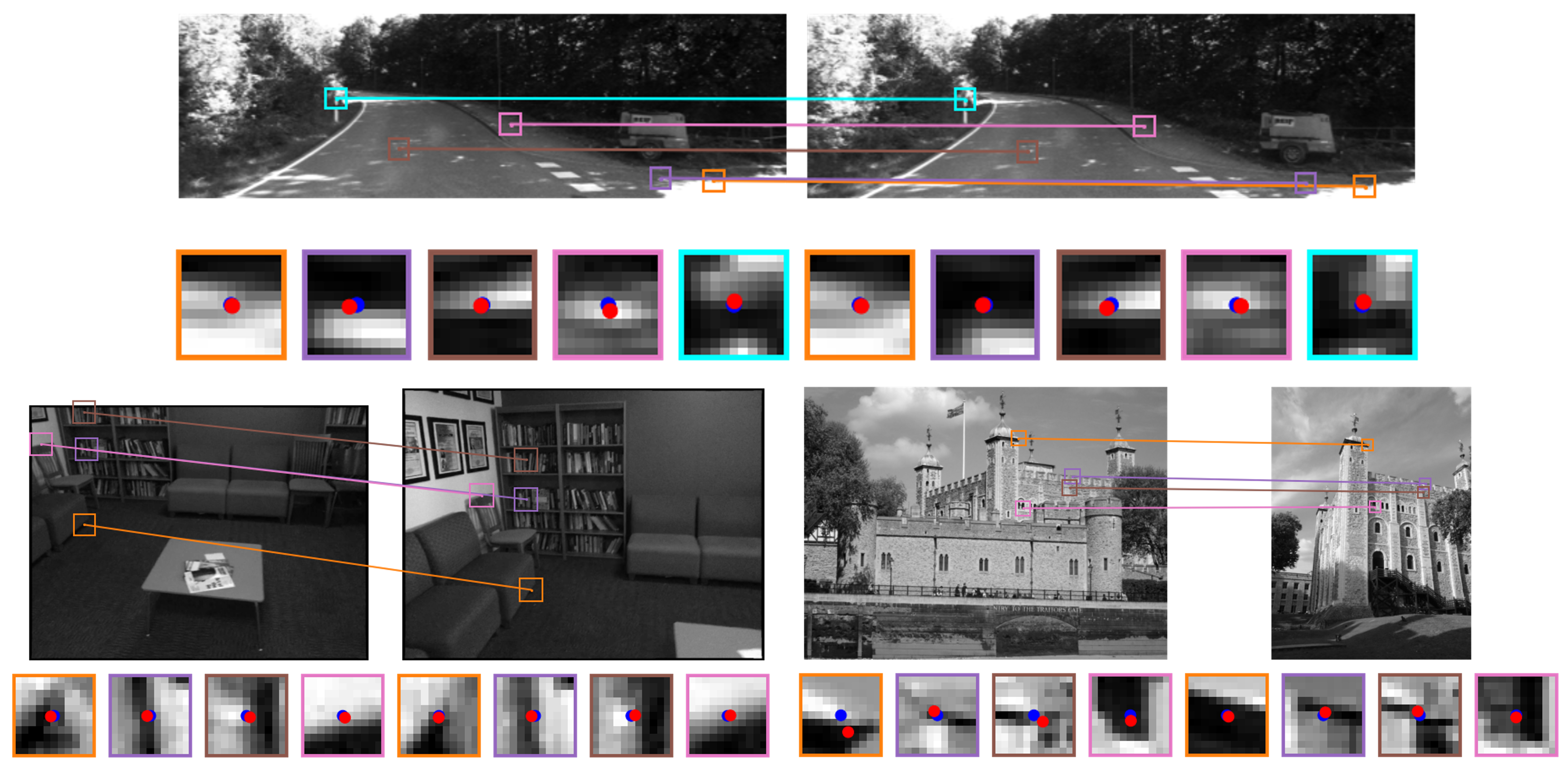}
  \caption{Image pairs from KITTI (top), ScanNet (bottom left), and MegaDepth (bottom right) datasets showing inliers of refined matches. 
  For each pair, the upper row shows matches, and the bottom one the image patches our Keypoint Refinement module takes, with initial points as blue and refined ones as red. 
  The refinements are larger on the MegaDepth pair, as it is visible also in the analysis provided by Fig.~\ref{fig:histograms}.
  However, both on KITTI and ScanNet, the refined features seem to be better localized visually.
  }
  \label{fig:qualitative}
\end{figure}

\subsection{Keypoint Localization}
To show that our refined keypoints are more precise in general, rather than merely rearranged for the setup of epipolar geometry, we evaluated the keypoint localization performance of our module on the HPatches benchmark~\cite{balntas2017hpatches}. We follow the setup described in LoFTR~\cite{sun2021loftr} and LightGlue~\cite{lindenberger2023lightglue}. Our method improved the combination of SuperPoint and mutual nearest neighbor matching (SP+MNN) from 27\% to 33\% Mean Matching Accuracy (MMA) at 1px. 

\subsection{Computational Analysis}

To show that the proposed Keypoint Refinement module is not a computationally significant part of the whole pipeline, we measured the inference time (Table~\ref{tab:computation}.) and number of parameters (Table~\ref{tab:numparams}.) of each component of the relative pose estimation pipeline. The inference time is measured for an image pair ($2\times2048$ keypoints), averaged on the MegaDepth-1500 benchmark.

From Table~\ref{tab:computation}, note that the computational time is less than $5\%$ for both detectors used with the Mutual Nearest Neighbor (MNN) matching, and it is straightforward that the portion will be smaller when used with the LightGlue matcher. From Table~\ref{tab:numparams}, we chose LightGlue matcher, and it takes the most significant portion of the learnable parameters, while ours only takes around $2\%$. Even if LightGlue is replaced with MNN that has no parameters, ours only takes $17\%$ of the detector.

\begin{table}[tb]
\renewcommand{\arraystretch}{1.1}
  \caption{The computation time of each component of the relative pose estimation pipeline (Fig.~\ref{fig:pipeline}). The first number of each cell represents the inference time taken on the component for 4096 keypoints (2048 matches) in milliseconds, and the parenthesized second number shows the portion of the inference time compared to the entire pipeline.}
  \label{tab:computation}
  \centering
  \resizebox{0.95\textwidth}{!}{\begin{tabular}{l|c c c | c}
    \toprule
    Method & Detection\&Matching & Ours  & Pose Estimation & Total\\ 
    \midrule
    SP + MNN + Ours & \phantom{1}$72$ms (37\%) & $7$ms (4\%) & $116$ms (60\%) & $195$ms \\
    ALIKED + MNN + Ours  & $163$ms (51\%) & $6$ms (2\%) & $150$ms (47\%) & $319$ms\\
  \bottomrule
  \end{tabular}}
  \vspace{-0.125cm}
\end{table}

\begin{table}[tb]
\renewcommand{\arraystretch}{1.1}
  \caption{The number of parameters possessed by each component of the relative pose estimation pipeline (Fig.~\ref{fig:pipeline}). The first number of each cell represents the number of parameters possessed by the component in millions, and the parenthesized second number shows the portion of the parameters compared to the entire pipeline.}
  \label{tab:numparams}
  \centering
  \resizebox{0.8\textwidth}{!}{\begin{tabular}{l|c c c | c}
    \toprule
    Method & Detection & Matching & Ours & Total\\ 
    \midrule
    SP+LG+Ours & 1.30M (10\%) & 11.85M (89\%) & 0.20M (2\%) & 13.35M \\
    ALIKED+LG+Ours & 0.68M (\phantom{1}5\%) & 11.85M (94\%) & 0.12M (1\%) & 12.65M \\
  \bottomrule
  \end{tabular}}
  \vspace{-0.175cm}
\end{table}
\subsection{Offset Analysis}

To decipher the learning mechanisms and the characteristics of offset vectors within the proposed network, we delve into the visualized scale and orientation histograms of the offsets predicted from SuperPoint keypoints, as illustrated across all tested datasets in Fig.~\ref{fig:histograms}. 
The upper part of the figures delineates the length distributions of the predicted displacement vectors, revealing pronounced displacements within the MegaDepth dataset. 
This observation aligns with the notable enhancements achieved on MegaDepth, implying a substantial potential for accuracy refinement through sub-pixel adjustments due to the initial imprecision in keypoint localization.

A progressive transition from left to right within these histograms demonstrates a trend towards diminished displacement lengths, indicating either an overall reduction in the necessity for keypoint adjustments or insufficient generalizability of our model similar to what neural-network based detectors have struggled.

Conversely, the lower section of Fig.~\ref{fig:histograms} presents the directional tendencies of these predicted displacements, showcasing a relatively uniform distribution with a predominant alignment along the vertical image axis. 
A notable deviation is observed along the horizontal axis, with a bias towards rightward adjustments. 
This pattern likely reflects dataset-specific attributes, suggesting that the observed directional imbalances may be influenced by the inherent visual structure and orientation within the datasets.

Qualitative examples are shown in Fig.~\ref{fig:qualitative}.
Here, image pairs from KITTI (top), ScanNet (bottom left), and MegaDepth (bottom right) datasets are visualized together with their original and refined inlier matches. 
For each pair, the upper row shows matches overlayed in the image pair.
The bottom one visualizes independent image patches our Keypoint Refinement module takes, with initial points as blue and refined ones as red.
The refinements are visibly larger on the image pair taken from the MegaDepth dataset.
However, both on KITTI and ScanNet, the refined features seem to be better localized.




\section{Conclusion}

This work introduces a novel post-processing procedure designed to endow any learned feature detection model with sub-pixel accuracy, addressing a critical limitation of recent detectors. 
By appending offset vectors to detected features, our method eliminates the need for designing specialized detectors while directly minimizing test-time evaluation metrics like relative pose error. 
Through rigorous testing on large-scale and real-world datasets with SuperPoint and ALIKED features and LightGlue matcher, we demonstrate consistent improvements in keypoint localization accuracy.
This is shown through relative pose and fundamental matrix estimation. 
Although the performance improvements are not markedly large, they are \textit{consistent} across various experiments, underscoring the reliability of our approach. 
Given the minimal computational overhead of our method, which adds only approx.\ $7$ milliseconds to the processing time, it offers a compelling improvement with no discernible drawbacks. 

\setcounter{section}{0}
\renewcommand\thesection{\Alph{section}}

\section{Ablation Study}
This section evaluates the effects of integrating SoftArgMax (SAM), Convolutional Neural Network (CNN), Descriptor Guidance (DG, $S_{ki}[u]=F_{ki}[u]d_{ki}$), and Mean Feature (MF, $S_{ki}[u]=F_{ki}[u]\cdot \frac{d_{1i}+d_{2i}}{2}$) to enhance feature detection on the MegaDepth dataset~\cite{li2018megadepth}, using SuperPoint features~\cite{detone2018superpoint} matched by mutual nearest neighbors (MNN).

Outlined in Table~\ref{tab:ablation}, the analysis begins with the performance of the vanilla SuperPoint baseline. 
The findings, detailed from the second row onwards, reveal that simply layering SAM atop SuperPoint does not markedly improve performance, underscoring SAM's limited standalone efficacy. 
Subsequent rows meticulously demonstrate how each proposed module—SAM, CNN, DG, and MF—individually contributes to the enhancement of detection accuracy. 
The culmination of these improvements is most notable when all modules are applied in tandem, resulting in the optimal enhancement of feature detection capabilities, as evidenced in the later sections of Table~\ref{tab:ablation}. 
This collective application underscores the potential of these enhancements in refining feature detectors on challenging datasets.

\section{Various Detector-Matcher Combinations}
In this section, we will further demonstrate the generalizability of our method on various detectors and matchers. 
We compare the performance of our method with the reinforced learning-based refinement~\cite{bhowmik2020reinforced} on the combination of SuperPoint~\cite{detone2018superpoint} and LightGlue~\cite{lindenberger2023lightglue}. 
Our method is also applied to the recent DeDoDe~\cite{edstedt2023dedode}, XFeat~\cite{potje2024xfeat}, Key.Net~\cite{barroso2019key}, and R2D2~\cite{revaud2019r2d2} to show further generalizability to state-of-the-art methods extracting local features. 
To discuss the limitation, we apply our method on top of SIFT~\cite{Lowe2004}. The results of this section are provided in Table~\ref{tab:splgresults}.

When applied on top of SuperPoint and LightGlue, the reinforcement learning approach~\cite{bhowmik2020reinforced} improves marginally, while the amount improved by our method is more significant. 
Note that the probabilistic feature matching of \cite{bhowmik2020reinforced} based on the softmax of Euclidean distance is not applicable without change when used with LightGlue. 
We instead modified it to use the soft assignment of LightGlue as the probability of a match to be sampled.

When our method is applied on top of SIFT, the performance drops slightly. 
This is the expected behaviour as the Difference of Gaussians (DoG) detector employed by SIFT, is already sub-pixel accurate. 
Thus there is no room to improve with the proposed method. 
In SIFT experiments, we increased the number of iterations of GC-RANSAC\cite{barath2018graph} to 5000, to maintain the stability of the robust estimator with worse initial matches. This disentangles the effect of our method from that of the stochasticity of the robust estimator.

To further demonstrate the generalizability of the proposed method on various detectors, we applied our method on one of the most recent state-of-the-art detectors, DeDoDe\cite{edstedt2023dedode, edstedt2024dedode}. Our method is shown to consistently improve DeDoDe on the provided metrics, even on the ScanNet\cite{dai2017scannet} benchmark on which the original detector severely fails. Note that DeDoDe utilizes their dual softmax matcher, denoted as DSM. We utilized the DeDoDe-L, DeDoDev2-L detectors and DeDoDe-G descriptor among their pre-trained models.

Another state-of-the-art detector, XFeat~\cite{potje2024xfeat}, can be improved with the proposed method. This is a straightforward result as its Keypoint Head resembles the structure of that of SuperPoint. Note that our module will not harm the strengths of XFeat, especially time efficiency, as ours are shown in the main paper to take around 7ms (which is approximately 143 image pairs per second) that is still insignificant compared to the inference time of XFeat on two images.

Additionally, we applied our method on top of two earlier detectors that aimed for sub-pixel accuracy (KeyNet~\cite{barroso2019key}) and that did not involve pooling layers (R2D2~\cite{revaud2019r2d2}), to show our method is not limited to make only keypoints sub-pixel accurate but even making already accurate keypoints to be more accurate in estimating relative poses.

\begin{table}[tb]
\renewcommand{\arraystretch}{1.1}
  \caption{\textbf{Ablation Study.} The impact of SoftArgMax (SAM), Convolutional Neural Network (CNN), Descriptor Guidance (DG), and Mean Feature (MF) on the MegaDepth dataset~\cite{li2018megadepth} is evaluated. 
  The baseline performance with vanilla SuperPoint (SP)~\cite{detone2018superpoint} is shown in the second row. 
  The third row reveals that simply adding SAM to SP is insufficient for significant improvement. 
  Each proposed module -- SAM, CNN, DG, and MF -- enhances accuracy individually, with the integration of all components resulting in the highest performance gains. The experiment on the second row with LightGlue (LG) matcher was conducted only on the validation set and showed similarly severe performance drop with what the second row with Mutual Nearest Neighbor (MNN) matcher showed. Confirming this failure case again on the test set does not have significantly different meaning from that of MNN matcher, and thus we skip to show the second row with LG matcher here.}
  \label{tab:ablation}
  \centering
  \resizebox{\textwidth}{!}{\begin{tabular}{@{}c|ccccc|cccccc@{}}
    \toprule
    Matcher & SP & +SAM & +CNN & +DG & +MF & AUC@$5^{\circ}$ & AUC@$10^{\circ}$ & AUC@$20^{\circ}$ & Inlier Ratio (\%) & Mean ($^{\circ}$) & Median ($^{\circ}$)\\
    \midrule
    \multirow{5}{*}{MNN} & \checkmark & &  &  &  & 35.34 & 45.37 & 54.22 & 33.78 & 27.89 & \phantom{1}3.94\\
     & \checkmark & \checkmark &  &  &  & 19.56 & 30.00 & 40.59 & 20.17 & 36.57 & 13.58\\
     & \checkmark & \checkmark & \checkmark &  &  & 36.71 & 47.52 & 56.59 & 33.75 & 28.09 & \phantom{1}3.98\\
     & \checkmark & \checkmark & \checkmark & \checkmark &  & 36.57 & 47.34 & 56.39 & 33.61 & 28.26 & \phantom{1}4.03\\
     & \checkmark & \checkmark & \checkmark & \checkmark & \checkmark & \textbf{37.16} & \textbf{48.07} & \textbf{57.15} & \textbf{34.42} & \textbf{27.69} & \phantom{1}\textbf{3.82}\\
    \midrule
    \multirow{5}{*}{LG} & \checkmark & &  &  &  & 59.82 & 72.88 & 82.31 & 79.34 & \phantom{1}7.39 & \phantom{1}1.25\\
     & \checkmark & \checkmark &  &  &  & - & - & - & - & - & -\\
     & \checkmark & \checkmark & \checkmark &  &  & 60.77 & 73.70 & 82.93 & 80.38 & \phantom{1}7.30 & \phantom{1}1.21\\
     & \checkmark & \checkmark & \checkmark & \checkmark &  & 60.76 & 73.68 & 80.27 & 80.26 & \phantom{1}7.32 & \phantom{1}1.21\\
    & \checkmark & \checkmark & \checkmark & \checkmark & \checkmark & \textbf{61.82} & \textbf{74.46} & \textbf{83.41} & \textbf{81.24} & \phantom{1}\textbf{7.22} & \phantom{1}\textbf{1.15}\\
  \bottomrule
  \end{tabular}}
\end{table}

\begin{table}[tb]
  \renewcommand{\arraystretch}{1.15}
  \caption{The relative pose estimation metrics evaluated with various combinations of local feature extractors and matchers. A detailed explanation of each metric is provided in our main paper.}
  \label{tab:splgresults}
  \centering
  \resizebox{\textwidth}{!}{\begin{tabular}{@{}c|l|cccccc@{}}
    \toprule
    Dataset & \multicolumn{1}{c|}{Method} & AUC@$5^{\circ}$ & AUC@$10^{\circ}$ & AUC@$20^{\circ}$ & Inlier Ratio (\%) & Mean ($^{\circ}$) & Median ($^{\circ}$)\\
    \midrule
     \multirow{15}{*}{MegaDepth} & SP+LG & 59.82 & 72.88 & 82.31 & 79.34 & \phantom{1}7.39 & \phantom{1}1.25\\
     & Reinforced SP+LG & 59.94 & 73.05 & 82.48 & 80.88 & \phantom{1}\textbf{7.12} & \phantom{1}1.25\\
     & SP+LG+Ours & \textbf{61.82} & \textbf{74.46} & \textbf{83.41} & \textbf{81.24} & \phantom{1}7.22 & \phantom{1}\textbf{1.15}\\
     \cmidrule{2-8}
     & SIFT+MNN & \phantom{1}\textbf{9.53} & \textbf{13.40} & \textbf{17.90} & \phantom{1}\textbf{5.90} & 74.85 & 65.84\\
     & SIFT+MNN+Ours & \phantom{1}9.49 & 13.34 & 17.84 & \phantom{1}5.89 & \textbf{74.77} & \textbf{65.74}\\
     \cmidrule{2-8}
     & DeDoDe+DSM & 48.83 & 61.70 & 71.76 & 78.08 & 16.26 & \phantom{1}1.93\\
     & DeDoDe+DSM+Ours & \textbf{52.57} & \textbf{65.15} & \textbf{74.68} & \textbf{81.94} & \textbf{15.06} & \phantom{1}\textbf{1.63}\\
     \cmidrule{2-8}
     & DeDoDev2+DSM & 55.49 & 67.37 & 76.23 & 82.66 & 14.73 & \phantom{1}1.41\\
     & DeDoDev2+DSM+Ours & \textbf{56.08} & \textbf{67.90} & \textbf{76.56} & \textbf{83.44} & \textbf{14.58} & \phantom{1}\textbf{1.38}\\
     \cmidrule{2-8}
     & Key.Net+MNN & 52.76 & 64.98 & 74.24 & 42.18 & 12.50 & \phantom{1}1.63\\
     & Key.Net+MNN+Ours & \textbf{53.78} & \textbf{65.83} & \textbf{74.92} & \textbf{42.86} & \textbf{12.15} & \phantom{1}\textbf{1.52}\\
     \cmidrule{2-8}
     & R2D2+MNN & 51.64 & 64.49 & 74.26 & 63.23 & 11.96 & \phantom{1}1.72\\
     & R2D2+MNN+Ours & \textbf{52.14} & \textbf{64.90} & \textbf{74.55} & \textbf{63.53} & \textbf{11.93} & \phantom{1}\textbf{1.71}\\
     \cmidrule{2-8}
     & XFeat+MNN & 41.37 & 52.88 & 62.42 & 33.83 & 18.38 & \phantom{1}2.85\\
     & XFeat+MNN+Ours & \textbf{43.11} & \textbf{54.40} & \textbf{63.62} & \textbf{35.15} & \textbf{17.92} & \phantom{1}\textbf{2.58}\\
     \midrule
    \multirow{4}{*}{KITTI} & SIFT+MNN & \textbf{78.23} & \textbf{88.07} & \textbf{93.71} & 25.15 & \phantom{1}\textbf{1.42} & \phantom{1}\textbf{0.65}\\
     & SIFT+MNN+Ours & 78.17 & 88.01 & 93.66 & \textbf{25.16} & \phantom{1}1.47 & \phantom{1}\textbf{0.65}\\
     \cmidrule{2-8}
     & DeDoDe+DSM & 71.21 & 83.01 & 90.10 & 67.28 & \phantom{1}\textbf{3.18} & \phantom{1}0.88\\
     & DeDoDe+DSM+Ours & \textbf{71.88} & \textbf{83.39} & \textbf{90.29} & \textbf{67.29} & \phantom{1}\textbf{3.18} & \phantom{1}\textbf{0.85}\\
     \midrule
    \multirow{4}{*}{ScanNet} & SIFT+MNN & \phantom{1}\textbf{1.69} & \phantom{1}\textbf{3.89} & \phantom{1}\textbf{7.32} & \phantom{1}\textbf{3.38} & 85.06 & 79.00\\
     & SIFT+MNN+Ours & \phantom{1}\textbf{1.69} & \phantom{1}3.77 & \phantom{1}7.13 & \phantom{1}3.37 & \textbf{84.56} & \textbf{78.46}\\
     \cmidrule{2-8}
     & DeDoDe+DSM & \phantom{1}3.18 & \phantom{1}6.34 & 10.16 & 89.32 & \textbf{131.2} & \textbf{180.0}\\
     & DeDoDe+DSM+Ours & \phantom{1}\textbf{3.25} & \phantom{1}\textbf{6.45} & \textbf{10.31} & \textbf{89.92} & \textbf{131.2} & \textbf{180.0}\\
  \bottomrule
  \end{tabular}}
\end{table}

\section{Analysis on Design Choice}
In this section, we show the effectiveness of the proposed method, by comparing it with other design choices for the keypoint refinement module and loss function. 

\subsection{Fine Matching: CNN vs. Transformer}
Recently, transformers have become the most popular neural network design. This trend has also been blended into the detector-free matchers, such as LoFTR~\cite{sun2021loftr}. On the other hand, our model does not utilize transformers in our pipeline despite its strong capability of finding relevance among all possible combinations, which is shown in LoFTR~\cite{sun2021loftr} to solve local correspondence refinement effectively. Here, we question the cost-efficiency of the transformer on the local correspondence refinement task, by comparing the performance of LoFTR's local refinement module and the proposed method.

As shown in Table~\ref{tab:oursvstf}, the relative pose estimation performance did not improve significantly when we replaced our model (except scale factor $\sigma=2.5$ and CNN; the fine module of LoFTR still needs initial features to be extracted) with the fine module of LoFTR. However, the fine module of LoFTR takes significantly more trainable parameters and inference time. Although this comparison is not completely valid as each of them was trained with its loss, it still supports that using transformers for a simple problem without thorough consideration can be suboptimal.

\begin{table}[tb]
    \caption{Experiments showing the advantage of our method against LoFTR's fine module~\cite{sun2021loftr}. The relative pose estimation metrics evaluated for naive CNN-based displacement estimation (SAM+CNN, 2nd row of Table~\ref{tab:ablation}), ours, and when LoFTR's fine module is attached upon ours (LoFTR Fine). Compared to LoFTR's fine module, our method performs significantly efficiently in both model size and inference time while only marginally worse for relative pose estimation.}
    \label{tab:oursvstf}
    \centering
    \setlength{\tabcolsep}{4pt}
    \resizebox{0.99\linewidth}{!}{\begin{tabular}{l|c c c c c c | c c}  
    \toprule
Method & AUC@$5^{\circ}$ & @$10^{\circ}$ & @$20^{\circ}$ & Inl.\ Rat.\ (\%) & Mean ($^{\circ}$) & Med.\ ($^{\circ}$) & Param. & Time\\ 
   \midrule
(SP+)SAM+CNN & 36.7 & 47.5 & 56.6 & 33.8 & 28.1 & 4.0 & \textbf{0.2M} & \textbf{7ms}\\
Ours & 37.2 & 48.1 & 57.2 & 34.4 & 27.7 & 3.8 & \textbf{0.2M} & \textbf{7ms}\\
LoFTR Fine & \textbf{37.4} & \textbf{48.4} & \textbf{57.4} & \textbf{34.6} & \textbf{27.5} & \textbf{3.7} & 1.5M & 58ms\\
\bottomrule
\end{tabular}}
\end{table}

\subsection{Loss Functions} \label{lossfnct}
In this section, we compare different loss functions with relative pose supervision to validate our choice of the epipolar loss function as shown in Table~\ref{tab:losses}. Among losses taking supervision from relative poses, LoFTR's and reprojection losses give marginally different performance. However, LoFTR's loss requires depth supervision. Also, we found triangulation needed for reprojection error numerically unstable for training.

\begin{table}[tb]
    \caption{Experiments showing the performance of relative pose estimation when the proposed keypoint refinement module is trained with various losses. Among those taking ground truth relative pose supervision, we compare two representative losses\textemdash epipolar and reprojection loss\textemdash and a recent loss proposed from a detector-free matcher, LoFTR~\cite{sun2021loftr}. Their performance does not significantly differ; however, the epipolar loss is selected based on its advantages mentioned in the \cref{lossfnct}.}
    \label{tab:losses}
    \centering
    \setlength{\tabcolsep}{4pt}
    \resizebox{0.99\linewidth}{!}{\begin{tabular}{l|c c c c c c }  
    \toprule
Loss & AUC@$5^{\circ}$ & @$10^{\circ}$ & @$20^{\circ}$ & Inl.\ Rat.\ (\%) & Mean ($^{\circ}$) & Med.\ ($^{\circ}$)\\ 
   \midrule
Epipolar (Ours) & 37.2 & \textbf{48.1} & \textbf{57.2} & 34.4 & 27.7 & \textbf{3.8}  \\
LoFTR's & \textbf{37.3} & \textbf{48.1} & \textbf{57.2} & \textbf{34.5} & \textbf{27.6} & \textbf{3.8}\\
Reprojection & 37.1 & 48.0 & 57.1 & 34.4 & 27.7 & 3.9\\
\bottomrule
\end{tabular}}
\vspace{-0.2cm}
\end{table}

\section{Experiment Details}
Images from MegaDepth dataset are resized to have a uniform longer dimension of 1024 pixels, preserving the original height-width ratio. Images of the KITTI and ScanNet datasets are resized to 1240$\times$376 and 640$\times$480 pixels, respectively.

For SuperPoint feature extractor, we used the same NMS radius and detection threshold with Reinforced SuperPoint~\cite{bhowmik2020reinforced} to make our comparison valid. The maximum number of keypoints extracted from each image is set to 2048 throughout all datasets and methods. For ALIKED feature extractor and LightGlue matcher, the default setting of the Glue Factory library~\cite{pautrat_suarez_2023_gluestick, lindenberger2023lightglue} is employed, except the aforementioned number of keypoints. For the SIFT detector and RootSIFT descriptor, we utilized OpenCV's implementation. For the results with extra detectors provided in the Table~\ref{tab:splgresults}, we utilized their official implementations without change.

\section{Additional Qualitative Examples}
\begin{figure}[tb]
  \centering
  \includegraphics[width=0.8\linewidth]{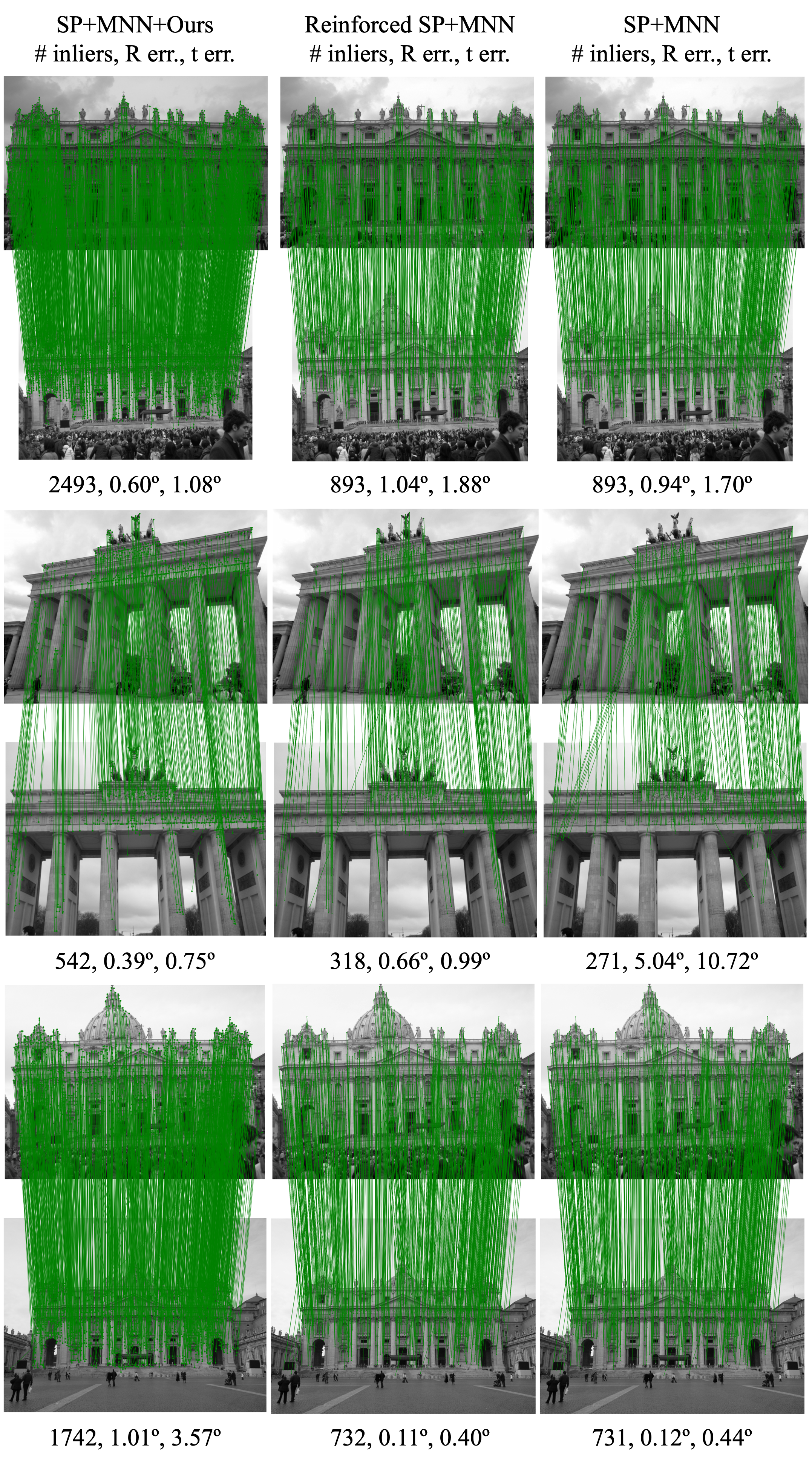}
  \caption{Visualization of geometric inlier correspondences on MegaDepth-1500 benchmark\cite{sun2021loftr} with our method, Reinforced SuperPoint, and SuperPoint, from left to right. The first and second row shows the successful cases and the last row shows the failure case. Note that our method consistently increases the number of inliers, but it does not always lead to more accurate relative poses.}
  \label{fig:match_megadepth}
\end{figure}

\begin{figure}[tb]
  \centering
  \includegraphics[width=0.75\linewidth]{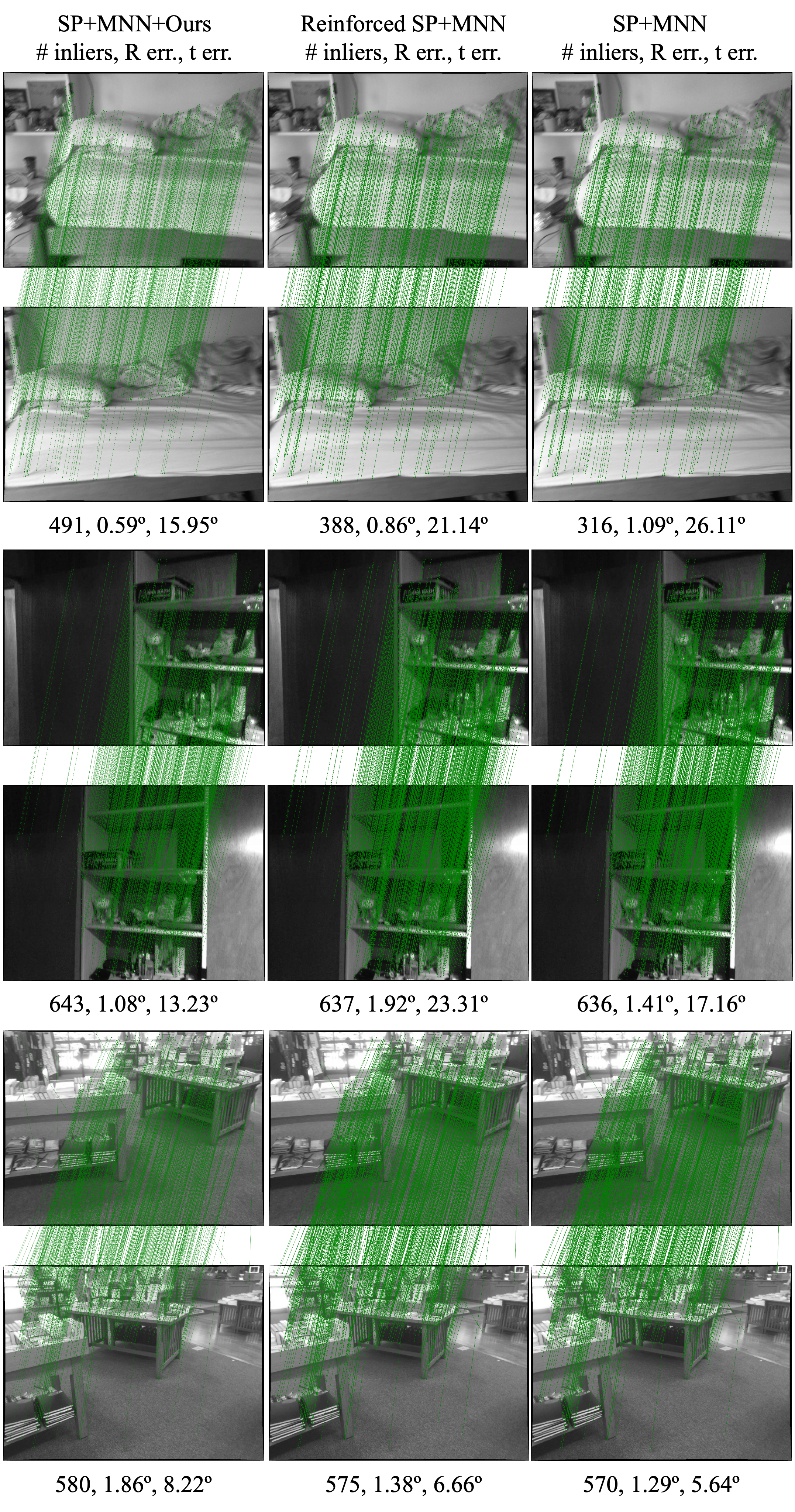}
  \caption{Visualization of geometric inlier correspondences on ScanNet benchmark\cite{sarlin2020superglue} with our method, Reinforced SuperPoint, and SuperPoint, from left to right. The first and second row shows the successful cases and the last row shows the failure case. Note that our method consistently increases the number of inliers, but it does not always lead to more accurate relative poses.}
  \label{fig:match_scannet}
\end{figure}

\begin{figure}[tb]
  \centering
  \includegraphics[width=1.0\linewidth]{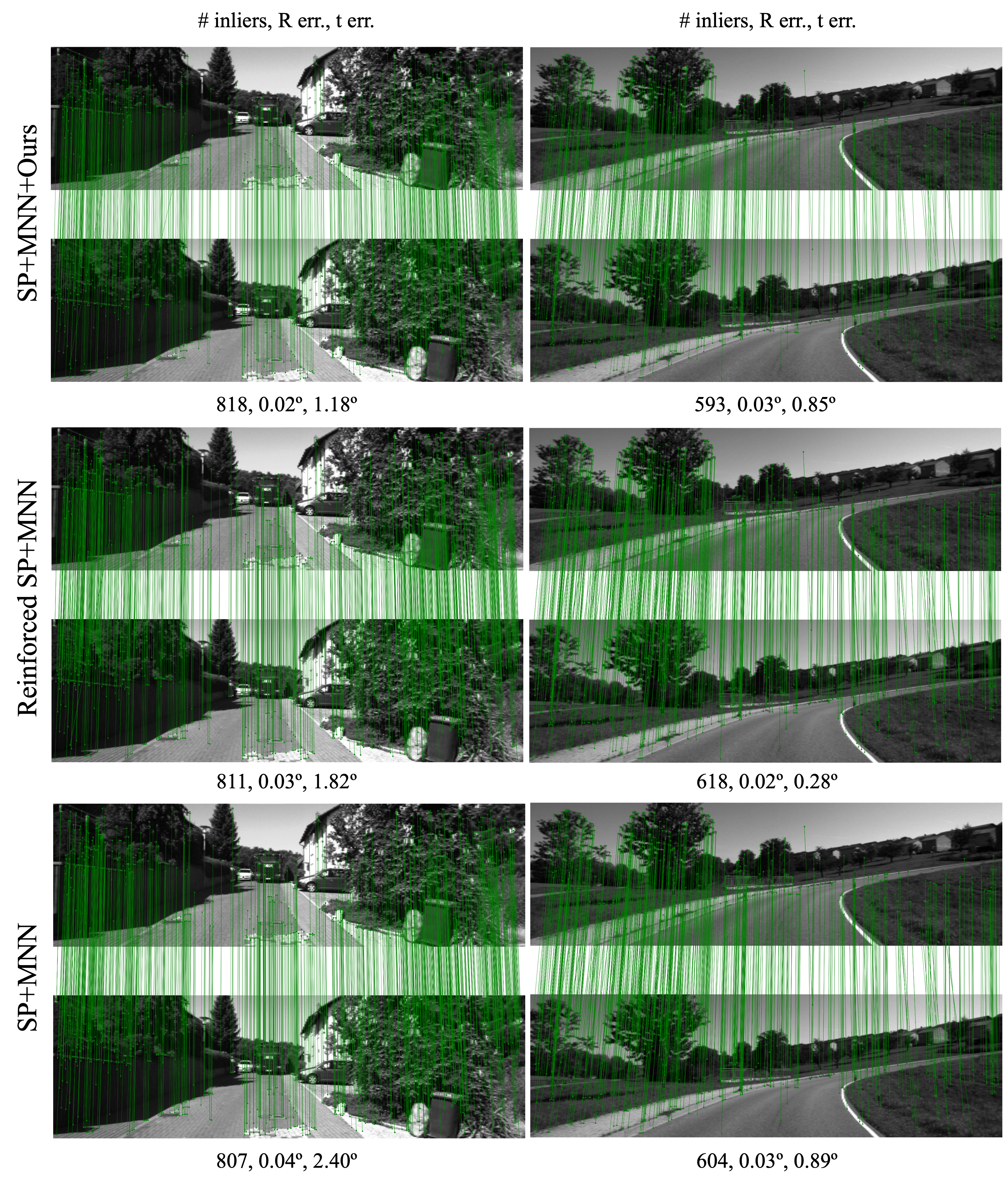}
  \caption{Visualization of geometric inlier correspondences on consecutive frames of scene 09 and 10 of the KITTI odometry dataset (following the data preprocessing procedure of\cite{2020_jau_zhu_deepFEPE}) with our method, Reinforced SuperPoint, and SuperPoint, from top to bottom. The left column shows the successful cases and the right column shows the failure case. Note that the performance differences between the methods are not very huge, as this is a relatively simple dataset; i.e. the viewpoint and illumination changes are insignificant along this dataset.}
  \label{fig:match_kitti}
\end{figure}

In this section, we visualize a few image pairs that are our successful and failed cases of refining the relative pose estimation, with the inlier matches used for the relative pose estimation. 
Each of \cref{fig:match_megadepth,fig:match_scannet,fig:match_kitti} shows the image pairs along with green lines, which are inlier correspondences used for relative pose estimation. 
The caption below each pair shows the number of inlier matches, rotation error, and translation error (degree) of the relative pose estimation. A detailed explanation of each metric is provided in our main paper.

\section*{Acknowledgement}

This work is conducted as a part of the \textit{Research in Computer Science} course at ETH Zurich, utilizing the computational resources of the ETH Computer Vision and Geometry (CVG) Group. The authors would like to thank Tong Wei, Alan Paul, and Fengshi Zheng for their helpful discussions.
This work was partially funded by the ETH RobotX research grant, and the Hasler Stiftung Research Grant via the ETH Zurich
Foundation.
\vfill
\pagebreak

%
%
\bibliographystyle{splncs04}
\bibliography{main_supp_arxiv}

\begin{thebibliography}{10}
\providecommand{\url}[1]{\texttt{#1}}
\providecommand{\urlprefix}{URL }
\providecommand{\doi}[1]{https://doi.org/#1}

\bibitem{Alahi2012}
Alahi, A., Ortiz, R., Vandergheynst, P.: Freak: Fast retina keypoint. In: 2012 IEEE conference on computer vision and pattern recognition. pp. 510--517. IEEE (2012)

\bibitem{arandjelovic2012three}
Arandjelovi{\'c}, R., Zisserman, A.: Three things everyone should know to improve object retrieval. In: 2012 IEEE conference on computer vision and pattern recognition. pp. 2911--2918. IEEE (2012)

\bibitem{arandjelovic2013all}
Arandjelovic, R., Zisserman, A.: All about vlad. In: Proceedings of the IEEE conference on Computer Vision and Pattern Recognition. pp. 1578--1585 (2013)

\bibitem{Balntas2016}
Balntas, V., Riba, E., Ponsa, D., Mikolajczyk, K.: Learning local feature descriptors with triplets and shallow convolutional neural networks. In: BMVC. vol.~1 (2016)

\bibitem{balntas2017hpatches}
Balntas, V., Lenc, K., Vedaldi, A., Mikolajczyk, K.: Hpatches: A benchmark and evaluation of handcrafted and learned local descriptors. In: Proceedings of the IEEE conference on computer vision and pattern recognition. pp. 5173--5182 (2017)

\bibitem{barath2018graph}
Barath, D., Matas, J.: Graph-cut ransac. In: Proceedings of the IEEE conference on computer vision and pattern recognition. pp. 6733--6741 (2018)

\bibitem{barath2020magsac++}
Barath, D., Noskova, J., Ivashechkin, M., Matas, J.: Magsac++, a fast, reliable and accurate robust estimator. In: Proceedings of the IEEE/CVF conference on computer vision and pattern recognition. pp. 1304--1312 (2020)

\bibitem{Barroso-Laguna2020}
Barroso-Laguna, A., Verdie, Y., Busam, B., Mikolajczyk, K.: Hddnet: Hybrid detector descriptor with mutual interactive learning. In: Proceedings of the Asian Conference on Computer Vision (November 2020)

\bibitem{barroso2019key}
Barroso-Laguna, A., Riba, E., Ponsa, D., Mikolajczyk, K.: Key. net: Keypoint detection by handcrafted and learned cnn filters. In: Proceedings of the IEEE/CVF international conference on computer vision. pp. 5836--5844 (2019)

\bibitem{Bay2008}
Bay, H., Ess, A., Tuytelaars, T., Van~Gool, L.: Speeded-up robust features (surf). Computer vision and image understanding  \textbf{110}(3),  346--359 (2008)

\bibitem{bhowmik2020reinforced}
Bhowmik, A., Gumhold, S., Rother, C., Brachmann, E.: Reinforced feature points: Optimizing feature detection and description for a high-level task. In: Proceedings of the IEEE/CVF conference on computer vision and pattern recognition. pp. 4948--4957 (2020)

\bibitem{brachmann2019neural}
Brachmann, E., Rother, C.: Neural-guided ransac: Learning where to sample model hypotheses. In: Proceedings of the IEEE/CVF International Conference on Computer Vision. pp. 4322--4331 (2019)

\bibitem{Calonder2010}
Calonder, M., Lepetit, V., Strecha, C., Fua, P.: Brief: Binary robust independent elementary features. In: European conference on computer vision. pp. 778--792. Springer (2010)

\bibitem{Choy2016}
Choy, C.B., Gwak, J., Savarese, S., Chandraker, M.: Universal correspondence network. Advances in neural information processing systems  \textbf{29} (2016)

\bibitem{chum2005matching}
Chum, O., Matas, J.: Matching with prosac-progressive sample consensus. In: 2005 IEEE computer society conference on computer vision and pattern recognition (CVPR'05). vol.~1, pp. 220--226. IEEE (2005)

\bibitem{chum2008optimal}
Chum, O., Matas, J.: Optimal randomized ransac. IEEE Transactions on Pattern Analysis and Machine Intelligence  \textbf{30}(8),  1472--1482 (2008)

\bibitem{chum2003locally}
Chum, O., Matas, J., Kittler, J.: Locally optimized ransac. In: Pattern Recognition: 25th DAGM Symposium, Magdeburg, Germany, September 10-12, 2003. Proceedings 25. pp. 236--243. Springer (2003)

\bibitem{dai2017scannet}
Dai, A., Chang, A.X., Savva, M., Halber, M., Funkhouser, T., Nie{\ss}ner, M.: Scannet: Richly-annotated 3d reconstructions of indoor scenes. In: Computer Vision and Pattern Recognition. pp. 5828--5839 (2017)

\bibitem{Dalal2005}
Dalal, N., Triggs, B.: Histograms of oriented gradients for human detection. In: 2005 IEEE computer society conference on computer vision and pattern recognition (CVPR'05). vol.~1, pp. 886--893. IEEE (2005)

\bibitem{detone2018superpoint}
DeTone, D., Malisiewicz, T., Rabinovich, A.: Superpoint: Self-supervised interest point detection and description. In: Proceedings of the IEEE conference on computer vision and pattern recognition workshops. pp. 224--236 (2018)

\bibitem{D2Net2019}
Dusmanu, M., Rocco, I., Pajdla, T., Pollefeys, M., Sivic, J., Torii, A., Sattler, T.: {D2-Net: A Trainable CNN for Joint Detection and Description of Local Features}. In: CVPR (2019)

\bibitem{edstedt2023dedode}
Edstedt, J., B{\"o}kman, G., Wadenb{\"a}ck, M., Felsberg, M.: Dedode: Detect, don't describe--describe, don't detect for local feature matching. arXiv preprint arXiv:2308.08479  (2023)

\bibitem{edstedt2024dedode}
Edstedt, J., B{\"o}kman, G., Zhao, Z.: Dedode v2: Analyzing and improving the dedode keypoint detector. In: Proceedings of the IEEE/CVF Conference on Computer Vision and Pattern Recognition. pp. 4245--4253 (2024)

\bibitem{Geiger2012CVPR}
Geiger, A., Lenz, P., Urtasun, R.: Are we ready for autonomous driving? the kitti vision benchmark suite. In: Conference on Computer Vision and Pattern Recognition (CVPR) (2012)

\bibitem{Han2015}
Han, X., Leung, T., Jia, Y., Sukthankar, R., Berg, A.C.: Matchnet: Unifying feature and metric learning for patch-based matching. In: Proceedings of the IEEE Conference on Computer Vision and Pattern Recognition. pp. 3279--3286 (2015)

\bibitem{Harris1988}
Harris, C., Stephens, M., et~al.: A combined corner and edge detector. In: Alvey vision conference. vol.~15, pp. 10--5244. Manchester, UK (1988)

\bibitem{hartley2003multiple}
Hartley, R., Zisserman, A.: Multiple view geometry in computer vision. Cambridge university press (2003)

\bibitem{Jaderberg2015}
Jaderberg, M., Simonyan, K., Zisserman, A., et~al.: Spatial transformer networks. Advances in neural information processing systems  \textbf{28} (2015)

\bibitem{jared2015reconstructing}
Jared, H., Schonberger, J.L., Dunn, E., Frahm, J.M.: Reconstructing the world in six days. In: CVPR (2015)

\bibitem{2020_jau_zhu_deepFEPE}
{Jau}, Y.Y., {Zhu}, R., {Su}, H., {Chandraker}, M.: Deep keypoint-based camera pose estimation with geometric constraints. In: 2020 IEEE/RSJ International Conference on Intelligent Robots and Systems (IROS). pp. 4950--4957 (2020). \doi{10.1109/IROS45743.2020.9341229}

\bibitem{IMC2020}
Jin, Y., Mishkin, D., Mishchuk, A., Matas, J., Fua, P., Yi, K.M., Trulls, E.: Image matching across wide baselines: From paper to practice. International Journal of Computer Vision  (2020)

\bibitem{lee2022self}
Jongmin~Lee, Byungjin~Kim, M.C.: Self-supervised equivariant learning for oriented keypoint detection. In: Proceedings of IEEE/CVF Conference on Computer Vision and Pattern Recognition (CVPR) (2022)

\bibitem{Lee_2023_CVPR}
Lee, J., Kim, B., Kim, S., Cho, M.: Learning rotation-equivariant features for visual correspondence. In: Proceedings of the IEEE/CVF Conference on Computer Vision and Pattern Recognition (CVPR) (June 2023)

\bibitem{Leutenegger2011}
Leutenegger, S., Chli, M., Siegwart, R.Y.: Brisk: Binary robust invariant scalable keypoints. In: 2011 International conference on computer vision. pp. 2548--2555. IEEE (2011)

\bibitem{li2012worldwide}
Li, Y., Snavely, N., Huttenlocher, D., Fua, P.: Worldwide pose estimation using 3d point clouds. In: European conference on computer vision. pp. 15--29. Springer (2012)

\bibitem{li2010location}
Li, Y., Snavely, N., Huttenlocher, D.P.: Location recognition using prioritized feature matching. In: Computer Vision--ECCV 2010: 11th European Conference on Computer Vision, Heraklion, Crete, Greece, September 5-11, 2010, Proceedings, Part II 11. pp. 791--804. Springer (2010)

\bibitem{li2018megadepth}
Li, Z., Snavely, N.: Megadepth: Learning single-view depth prediction from internet photos. In: Proceedings of the IEEE conference on computer vision and pattern recognition. pp. 2041--2050 (2018)

\bibitem{lindenberger2021pixsfm}
Lindenberger, P., Sarlin, P.E., Larsson, V., Pollefeys, M.: {Pixel-Perfect Structure-from-Motion with Featuremetric Refinement}. In: ICCV (2021)

\bibitem{lindenberger2023lightglue}
Lindenberger, P., Sarlin, P.E., Pollefeys, M.: Lightglue: Local feature matching at light speed. ICCV  (2023)

\bibitem{Liu2019}
Liu, Y., Shen, Z., Lin, Z., Peng, S., Bao, H., Zhou, X.: Gift: Learning transformation-invariant dense visual descriptors via group cnns. Advances in Neural Information Processing Systems  \textbf{32} (2019)

\bibitem{Lowe2004}
Lowe, D.G.: Distinctive image features from scale-invariant keypoints. International journal of computer vision  \textbf{60}(2),  91--110 (2004)

\bibitem{Luo2020}
Luo, Z., Zhou, L., Bai, X., Chen, H., Zhang, J., Yao, Y., Li, S., Fang, T., Quan, L.: {ASLFeat}: Learning local features of accurate shape and localization. In: Proceedings of the IEEE/CVF Conference on Computer Vision and Pattern Recognition (Apr 2020)

\bibitem{Mishchuk2018}
Mishchuk, A., Mishkin, D., Radenovic, F., Matas, J.: Working hard to know your neighbor’s margins: Local descriptor learning loss. In: Advances in Neural Information Processing Systems (Jan 2018)

\bibitem{Mishkin2018}
Mishkin, D., Radenovic, F., Matas, J.: Repeatability is not enough: Learning affine regions via discriminability. In: Proceedings of the European Conference on Computer Vision (ECCV). pp. 284--300 (2018)

\bibitem{more2006levenberg}
Mor{\'e}, J.J.: The levenberg-marquardt algorithm: implementation and theory. In: Numerical analysis: proceedings of the biennial Conference held at Dundee, June 28--July 1, 1977. pp. 105--116. Springer (2006)

\bibitem{Mur-Artal2015}
Mur-Artal, R., Montiel, J.M.M., Tardos, J.D.: Orb-slam: A versatile and accurate monocular slam system. IEEE Transactions on Robotics  \textbf{31},  1147--1163 (2015)

\bibitem{mur2017orb}
Mur-Artal, R., Tard{\'o}s, J.D.: Orb-slam2: An open-source slam system for monocular, stereo, and rgb-d cameras. IEEE transactions on robotics  \textbf{33}(5),  1255--1262 (2017)

\bibitem{nister2006scalable}
Nister, D., Stewenius, H.: Scalable recognition with a vocabulary tree. In: 2006 IEEE Computer Society Conference on Computer Vision and Pattern Recognition (CVPR'06). vol.~2, pp. 2161--2168. Ieee (2006)

\bibitem{noh2017large}
Noh, H., Araujo, A., Sim, J., Weyand, T., Han, B.: Large-scale image retrieval with attentive deep local features. In: Proceedings of the IEEE international conference on computer vision. pp. 3456--3465 (2017)

\bibitem{Ono2018}
Ono, Y., Trulls, E., Fua, P., Yi, K.M.: Lf-net: Learning local features from images. In: Advances in Neural Information Processing Systems 31. pp. 6234--6244. Curran Associates, Inc. (2018)

\bibitem{pautrat_suarez_2023_gluestick}
Pautrat*, R., Su{\'a}rez*, I., Yu, Y., Pollefeys, M., Larsson, V.: {GlueStick: Robust Image Matching by Sticking Points and Lines Together}. In: International Conference on Computer Vision (ICCV) (2023)

\bibitem{philbin2007object}
Philbin, J., Chum, O., Isard, M., Sivic, J., Zisserman, A.: Object retrieval with large vocabularies and fast spatial matching. In: 2007 IEEE conference on computer vision and pattern recognition. pp.~1--8. IEEE (2007)

\bibitem{potje2024xfeat}
Potje, G., Cadar, F., Araujo, A., Martins, R., Nascimento, E.R.: Xfeat: Accelerated features for lightweight image matching. In: Proceedings of the IEEE/CVF Conference on Computer Vision and Pattern Recognition. pp. 2682--2691 (2024)

\bibitem{Rao2022}
Rao, Y., Yang, J., Ju, Y., Li, C., Rigall, E., Fan, H., Dong, J.: Learning general feature descriptor for visual measurement with hierarchical view consistency. IEEE Transactions on Instrumentation and Measurement  \textbf{71},  1--12 (2022)

\bibitem{revaud2019r2d2}
Revaud, J., De~Souza, C., Humenberger, M., Weinzaepfel, P.: R2d2: Reliable and repeatable detector and descriptor. Advances in neural information processing systems  \textbf{32} (2019)

\bibitem{R2D22019}
Revaud, J., Weinzaepfel, P., De~Souza, C., Pion, N., Csurka, G., Cabon, Y., Humenberger, M.: R2d2: Repeatable and reliable detector and descriptor. In: NeurIPS (2019)

\bibitem{Roessle_2023_ICCV}
Roessle, B., Nie{\ss}ner, M.: End2end multi-view feature matching with differentiable pose optimization. In: Proceedings of the IEEE/CVF International Conference on Computer Vision (ICCV). pp. 477--487 (October 2023)

\bibitem{Rosten2006}
Rosten, E., Drummond, T.: Machine learning for high-speed corner detection. In: European conference on computer vision. pp. 430--443. Springer (2006)

\bibitem{Rublee2011}
Rublee, E., Rabaud, V., Konolige, K., Bradski, G.: Orb: An efficient alternative to sift or surf. In: 2011 International conference on computer vision. pp. 2564--2571. IEEE (2011)

\bibitem{sarlin2020superglue}
Sarlin, P.E., DeTone, D., Malisiewicz, T., Rabinovich, A.: Superglue: Learning feature matching with graph neural networks. In: Proceedings of the IEEE/CVF conference on computer vision and pattern recognition. pp. 4938--4947 (2020)

\bibitem{sattler2016efficient}
Sattler, T., Leibe, B., Kobbelt, L.: Efficient \& effective prioritized matching for large-scale image-based localization. IEEE transactions on pattern analysis and machine intelligence  \textbf{39}(9),  1744--1756 (2016)

\bibitem{schindler2007city}
Schindler, G., Brown, M., Szeliski, R.: City-scale location recognition. In: 2007 IEEE Conference on Computer Vision and Pattern Recognition. pp.~1--7. IEEE (2007)

\bibitem{Schonberger2016}
Sch{\"o}nberger, J.L., Frahm, J.M.: Structure-from-motion revisited. In: Proceedings of the IEEE conference on computer vision and pattern recognition. pp. 4104--4113 (2016)

\bibitem{sun2021loftr}
Sun, J., Shen, Z., Wang, Y., Bao, H., Zhou, X.: Loftr: Detector-free local feature matching with transformers. In: Proceedings of the IEEE/CVF conference on computer vision and pattern recognition. pp. 8922--8931 (2021)

\bibitem{Suwajanakorn2018}
Suwajanakorn, S., Snavely, N., Tompson, J.J., Norouzi, M.: Discovery of latent 3d keypoints via end-to-end geometric reasoning. In: NeurIPS (2018)

\bibitem{Suwanwimolkul2021}
Suwanwimolkul, S., Komorita, S., Tasaka, K.: Learning of low-level feature keypoints for accurate and robust detection. In: 2021 IEEE Winter Conference on Applications of Computer Vision (WACV). pp. 2261--2270. IEEE, Waikoloa, HI, USA (Jan 2021)

\bibitem{Tian2020}
Tian, Y., Balntas, V., Ng, T., Barroso-Laguna, A., Demiris, Y., Mikolajczyk, K.: {D2D}: Keypoint extraction with describe to detect approach. In: Proceedings of the Asian Conference on Computer Vision (2020)

\bibitem{Tian2017}
Tian, Y., Fan, B., Wu, F.: L2-net: Deep learning of discriminative patch descriptor in euclidean space. In: 2017 IEEE Conference on Computer Vision and Pattern Recognition. pp. 6128--6136. Honolulu, HI (Jul 2017)

\bibitem{Tian2019}
Tian, Y., Yu, X., Fan, B., Wu, F., Heijnen, H., Balntas, V.: Sosnet: Second order similarity regularization for local descriptor learning. In: Conference on Computer Vision and Pattern Recognition (Dec 2019)

\bibitem{toft2018semantic}
Toft, C., Stenborg, E., Hammarstrand, L., Brynte, L., Pollefeys, M., Sattler, T., Kahl, F.: Semantic match consistency for long-term visual localization. In: Proceedings of the European Conference on Computer Vision (ECCV). pp. 383--399 (2018)

\bibitem{DISK2020}
Tyszkiewicz, M.J., Fua, P., Trulls, E.: Disk: Learning local features with policy gradient. In: NeurIPS (2020)

\bibitem{wei2023generalized}
Wei, T., Patel, Y., Shekhovtsov, A., Matas, J., Barath, D.: Generalized differentiable ransac. In: Proceedings of the IEEE/CVF International Conference on Computer Vision. pp. 17649--17660 (2023)

\bibitem{wu2013towards}
Wu, C.: Towards linear-time incremental structure from motion. In: 2013 International Conference on 3D Vision-3DV 2013. pp. 127--134. IEEE (2013)

\bibitem{Yi2016}
Yi, K.M., Trulls, E., Lepetit, V., Fua, P.: Lift: Learned invariant feature transform. In: European Conference on Computer Vision. vol.~9910, pp. 467--483. Springer (2016)

\bibitem{yi2018learning}
Yi, K.M., Trulls, E., Ono, Y., Lepetit, V., Salzmann, M., Fua, P.: Learning to find good correspondences. In: Proceedings of the IEEE conference on computer vision and pattern recognition. pp. 2666--2674 (2018)

\bibitem{zhang2019learning}
Zhang, J., Sun, D., Luo, Z., Yao, A., Zhou, L., Shen, T., Chen, Y., Quan, L., Liao, H.: Learning two-view correspondences and geometry using order-aware network. In: Proceedings of the IEEE/CVF international conference on computer vision. pp. 5845--5854 (2019)

\bibitem{zhao2021progressive}
Zhao, C., Ge, Y., Zhu, F., Zhao, R., Li, H., Salzmann, M.: Progressive correspondence pruning by consensus learning. In: Proceedings of the IEEE/CVF International Conference on Computer Vision. pp. 6464--6473 (2021)

\bibitem{Zhao2023ALIKED}
Zhao, X., Wu, X., Chen, W., Chen, P.C.Y., Xu, Q., Li, Z.: {ALIKED}: A lighter keypoint and descriptor extraction network via deformable transformation. IEEE Transactions on Instrumentation \& Measurement  \textbf{72},  1--16 (2023). \doi{10.1109/TIM.2023.3271000}, \url{https://arxiv.org/pdf/2304.03608.pdf}

\bibitem{Zhao2022ALIKE}
Zhao, X., Wu, X., Miao, J., Chen, W., Chen, P.C.Y., Li, Z.: Alike: Accurate and lightweight keypoint detection and descriptor extraction. IEEE Transactions on Multimedia  (2022). \doi{10.1109/TMM.2022.3155927}, \url{http://arxiv.org/abs/2112.02906}

\end{thebibliography}
\end{document}